\documentclass[letterpaper,twocolumn,10pt]{article}
\usepackage{template}
\usepackage{comment}
\usepackage{tikz}
\usepackage{amsmath}
\usepackage{booktabs}
\usepackage{cleveref}

\usepackage{amssymb}

\DeclareMathOperator*{\argmin}{arg\,min}
\pdfoutput=1

\begin{document}
%-------------------------------------------------------------------------------

\date{}

\title{\Large \bf Multi-concept adversarial attacks} 

\author{
{\rm Vibha Belavadi}\\
The University of Texas at Dallas
\and
{\rm Yan Zhou}\\
The University of Texas at Dallas
\and
{\rm Murat Kantarcioglu}\\
The University of Texas at Dallas
\and
{\rm Bhavani M. Thuraisingham}\\
The University of Texas at Dallas
}

\maketitle

\begin{abstract}
%-------------------------------------------------------------------------------
As machine learning (ML) techniques are being 
increasingly used in many applications, their vulnerability to adversarial attacks becomes well-known. Test time attacks, typically launched by adding adversarial noise to test instances, have been shown effective against the deployed ML models. 
In practice, one test input may be leveraged by different ML models. For example, given an online profile image, one ML model could be used to classify the attractiveness of the individual, while the same profile image could be used by another model for age prediction. Test time attacks targeting a single ML model often neglect their impact on other ML models. In this work, we  empirically demonstrate that naively attacking the classifier learning one concept  may negatively impact classifiers trained to learn other concepts. For example, for the online image classification scenario, when the {\em Gender} classifier is under attack, the (wearing) {\em Glasses} classifier is simultaneously ``attacked'' with the accuracy  dropped from 98.69\% to 88.42\%. This raises an interesting research question: is it possible to
attack one set of classifiers without impacting the other set that uses the same test instance?

Answers to the above research question have  interesting implications for protecting privacy against ML model misuse. Attacking ML models that pose unnecessary risks of privacy invasion can be an important tool for protecting individuals from harmful privacy exploitation. For example, a discriminatory ML model that tries to predict the sexual orientation of the individuals using their profile image may be attacked without impacting the performance of other classifiers (e.g. smile detection) that use the same image.

In this paper, we address the above research question by developing novel attack techniques that can simultaneously
attack one set of ML models while preserving the accuracy of the other. In the case of linear classifiers, we provide a theoretical framework for finding an optimal solution to generate such  adversarial examples. Using this theoretical framework, we develop a ``multi-concept'' attack strategy in the context of deep learning tasks. Our results demonstrate that our techniques can successfully attack the target classes while protecting the ``protected'' classes in many different settings, which is not possible with the existing test-time attack-single strategies. 
\end{abstract}

\section{Introduction}
%-------------------------------------------------------------------------------
Machine learning models are being increasingly integrated into every aspect of modern life ranging from finance to healthcare. At the same time, recent work has shown that such ML models could be easily attacked. In the case of test time attacks, ML models could be fooled by test instances with  added adversarial noise. 
Although test-time attacks are shown to be effective, and do not require access to the training data, existing work usually considers only one ML model/classifier at a time. 
For example, an attacker who wants to attack a classifier that predicts {\em sexual orientation} from  social media images may choose to use  existing attack techniques (e.g., the ones against DNNs as in~\cite{madry2019deep}), and modify the facial images by adding a small amount of  adversarial noise to distort its embedded sexual orientation. 
Although attacking a single  classifier has been extensively studied, such test time attacks may have important side effects in many practical deployment settings, especially when the same instances are repeatedly exposed to and assessed by multiple ML models trained for different classification tasks.  
As we show in our experimental evaluation, using the existing test-time attack techniques would have undesirable side effects that, while attacking one classifier (e.g., predicting sexual orientation), may reduce the accuracy of another classifier (e.g., predicting attractiveness). 
This raises an interesting research question: \textit{Is it possible to launch test time attacks that only impact one set of  classifiers without impacting the other that uses the same test instance?}

Attacking one classifier while preserving the others 
has important implications for protecting privacy against discriminatory classifiers.
For example, an ML model may be trained to predict privacy-sensitive information such as sexual orientation and/or political affiliation using social network profiles~\cite{SMShadowProfileOrientation, PoliticalKosinski, OrientationKosinski}.
Recent work has shown similar intrusive usages of ML models, predicting attractiveness using profile pictures~\cite{DPicAttr, DPAttr_paper2}, and reporting life satisfaction~\cite{LifeSat_paper1, LifeSat_paper2}. %etc. 
Although some of these usages seem innocuous, %still 
the deployment of such ML models may be seen as a threat to individual privacy. \footnote{We would like to stress that in this case, the privacy challenge occurs because the ML model predicts sensitive information. Hence, privacy-preserving ML techniques that output accurate ML models will not address this privacy challenge.} Therefore, launching test time attack against one privacy violating model without impacting the performance of  other ML models could be beneficial for such scenarios (e.g., modifying the online profile image to look more attractive to one ML model without impact other models that use the same profile image).

To address the challenge of simultaneously attacking multiple ML models/classifiers while protecting the classification accuracy of other chosen classifiers, we propose a novel test-time attack algorithm. Unlike the existing test-time attacks, our algorithm considers the existence of the multiple classifiers and carefully adjusts the utility function to find an attack that considers multiple constraints (e.g., attack sexual orientation classifier while minimizing the impact on the attractiveness classifier, or attack both sexual orientation and attractiveness classifier, etc.).  For linear classifiers, we provide the optimal formalization of the problem and show how it could be solved using integer linear programming techniques. Later on, using this theoretical foundation, we provide an attack algorithm for Deep Neural Network models (DNNs) and show that it can reduce the side effect of the traditional attacks that do not consider these multiple attack constraints. 

Although our multi-concept attack algorithm is mainly motivated by privacy applications, the algorithm itself is of interest to many other applications, especially for settings where the application domain may mandate some constraints on the attacks. For example, consider an attack that modifies facial images to prevent detection of sexual orientation of a set of individuals with known gender values (e.g., when it is known that all the people in the test set are male).  We show that our techniques can produce an attack against several highly correlated concepts while having a minimal impact on others.  

Our contributions of this work can be summarized as follows:
\begin{itemize}
    \item We empirically show that the existing test-time attacks may have side effects (e.g., reduce the accuracy) on other classifiers that make use of the same attacked instance.
    \item We provide a theoretical formalization of our multi-concept attack problem for linear classifiers and provide optimal solutions using integer linear programming. 
    \item We show how to extend our theoretical model developed for linear classifiers to address similar challenges encountered in learning with DNNs by integrating multiple objectives into the attack generation problem.
    \item Using extensive empirical evaluation, we show that our multi-concept attack algorithm significantly improves the targeted attack performance compared to the existing test-time attack techniques.
    \item Using the Shapely value, an indicator of each feature's contribution to prediction, we illustrate features most influential to the success of attacks, and how 
    our attack model successfully preserves the important features of the protected concepts.
    
\end{itemize}

The rest of the paper is organized as follows. 

In Section~\ref{sec:related}, we discuss related work in the domain of adversarial examples, specifically the multi-concept classification scenario, and explain the motivation behind our work. 

In section~\ref{sec:method}, we propose the generic problem formulation and show it could be solved for linear and non-linear (e.g. deep neural networks) classifiers.

In Section~\ref{sec:exp}, we present the experimental results for selective attack against Linear and non-Linear Classifiers.
Section~\ref{sec:conclusion} concludes our work. 
%\end{itemize}

\section{Related Work}
\label{sec:related}
%-------------------------------------------------------------------------------

The term {\em adversarial example} and its impact on the accuracy/reliability of Deep Neural Networks (DNNs) were introduced formally in~\cite{Szegedy}.
Since then, there have been a wide variety of adversarial attacks and defenses~\cite{AdvxReview,AdvxReview2}. 
Adversarial training~\cite{43405, 35184, MultiLabelAdversarialTraining} was proposed as a way to produce more robust models. There were several works~\cite{dong2017interpretable, 7727230, tramer2017space} that discuss the interpretability of DNNs for adversarial examples.~\cite{AdvxReview} and~\cite{AdvxReview2} provide a recent summary of existing adversarial attack and defense methods. 

The work in~\cite{MultiLabelDomainKnow} takes user domain knowledge formulated in the form of First-Order Logic (FOL) constraints to improve adversarial robustness and help detect adversarial examples. This approach requires a super-class to create rejection criteria constraints. These constraints get enforced on the unlabelled data points during the learning process. This ensures that the decision boundaries created are in line with the actual marginal distribution. At testing time, these rejection criteria constraints also help in detecting and rejecting adversarial samples. The authors evaluated their approach on ANIMAL, CIFAR, and PASCAL datasets in which super-class and sub-class concepts are defined. 

In a multi-classification problem, each instance has a unique label from a set of possible classes. Given an unlabeled instance, a  classifier predicts one of the classes for a given instance. In~\cite{MultiLabelAdvSets}, the authors tackle the problem of adversarial robustness in an ensemble of multi-class classifier models.  They consider the following scenarios: 1) mis-classification of the same concept (e.g., digit 0) by all the models in the ensemble set, and 2) mis-classification of a given concept by one model but correct prediction by other models in the ensemble set. In our work, we also considers multiple ML models. However, each model  is trained to learn a different concept, and our goal is to attack some  concept (e.g. {\em gender}) while  protecting another(e.g., {\em age}).

To the best of our knowledge, unlike the related work presented in this section, our work is the first to consider the problem of attacking multiple classifiers trained to learn different concepts,  while protecting the remaining concepts without a significant degradation in the accuracy of the latter. Our perturbations in the $L_p$ norm formulation are also bounded by $\epsilon$ value, unlike the previous work where a successful multi-concept attack required large distortions. 

In the case where two  concepts are correlated, when one is attacked, the other is inevitably impacted. Even when concepts are not related, adding noise to an instance to defeat one classifier may nevertheless affect other classifiers. Hence, attacking some concepts while protecting others is of great importance in real applications where instances may have multiple occasion of usage by various ML models.

\section{Problem Definition}
\label{sec:method}
 
In a multi-concept classification setting, there's a one-to-many mapping between every instance and its concept set. Thus the attacks proposed in this setting target data sets on which multiple learning tasks can be performed. Each successfully learned concept is a potential victim. More interestingly, the objective of the ``attacker'' is often  to attack some while protecting others. One such example in the real world is ``modifying'' one's picture in such a way that one learning model will predict ``young'' while the other will predict ``professional'' with years of experience. Here the adversarial example has attacked the ``age'' concept while successfully preserving the ``professional'' concept. Directly applying state-of-art attacks such as PGD $L_2$ and PGD $L_\infty$ on a particular concept, say in this case ``age'', does not guarantee no degradation in the accuracy of another concept learned from the same dataset. We will further discuss this in our experimental section. Our goal is to generate adversarial examples attacking a set of concepts without causing a significant drop accuracy of our desired protected concept set. 

Let $X \sim \mathcal{D}$ be the data set from a distribution $\mathcal{D}$ on which a set of learning models is trained, resulting $F = \{f_1(X), f_2(X), \ldots, f_n(X)\}$ where $f_i(X)$ is a decision function learned from $X$. Let $x$ be an instance drawn randomly from $\mathcal{D}$, $y_{i\in[1,n]}$ be the label of $x$ for the $i^{th}$ learning task. The attacker's objective is to corrupt $x$ by $\Delta x$ so that $F_a = \bigcup_{j\in J}F_j \subset F$ is attacked while $F_p = \bigcup_{k\in K}F_k \subset F | F_p \cap F_a = \emptyset$ is protected, with a minimum cost. Formally,

\begin{align}
\begin{split}
%min  & \quad ||c \odot \Delta x|| \\
\min & \quad H(\Delta x) \\
s.t. 
    %& \quad y \cdot (\mathbf{w}_{k_1, \ldots, k_{\ell}} x' + b )  \le 0\\
     & \quad \forall k_i \in J, \quad f_{k_i} (x + \Delta x)  \ne y_{k_i}\\
     & \quad \forall k_j \in K, \quad f_{k_j} (x + \Delta x) = y_{k_j} 
%&\quad y \cdot (\mathbf{w}_{K-\{k_1, \ldots, k_{\ell}\}} x' + b )  \ge 0
\end{split}
\label{eq:general}
\end{align}
where $H$ is an objective function. $J$ and $K$ are the index sets of the attacked and the protect model sets respectively. 

Specific formulations of the problem defined in~(\ref{eq:general}) fall into two general types, depending on the nature of the learning problems. When the learning problem is linear, attacking/protecting multiple linear decision functions can be formulated as a linear programming problem. Assume binary classification for simplicity (although multi-class concepts can be learned with one-against-all binary linear classifiers). Some of the strategies that can be employed in the multi-class scenario are choosing the label with the largest loss derivative, loss grouping by correlation of labels, etc. We will be discussing multi-class formulation in future work.  Suppose we are given $m$ binary linear classifiers with coefficients denoted as $\mathbf{w}_{i \in [1,n]}$, the problem  defined in Eq.~(\ref{eq:general}) can be specified as finding the minimum-cost perturbation $\Delta x$ to a given input $(x,y_{i\in[1,n]})$, where $y_i \in \{-1, +1\}$ is the label of $x$ in the $i^{th}$ problem space, such that only a subset $J$ of the linear classifiers outputs a false prediction for $x'$ where $x' = x + \Delta x$:
\begin{align*}
min  & \quad ||c \odot \Delta x||_p \\
s.t. 
    %& \quad y \cdot (\mathbf{w}_{k_1, \ldots, k_{\ell}} x' + b )  \le 0\\
     & \quad \forall k_i \in J, \quad \quad \quad y_{k_i}\cdot [\mathbf{w}_{k_i} (x + \Delta x) + b_{k_i}]  \le 0\\
     & \quad \forall k_j \in [1,n]/J, \quad y_{k_j} \cdot [\mathbf{w}_{k_j} (x + \Delta x) + b_{k_j}]  > 0 
%&\quad y \cdot (\mathbf{w}_{K-\{k_1, \ldots, k_{\ell}\}} x' + b )  \ge 0
\end{align*}
where $c$ is the cost of modifying each feature, $\odot$ denotes element wise product, %and $||\cdot||$ is a distortion measure. $c$ is the cost to modify the attributes 
and $||\cdot||_p$ denotes $L_p$ norm. This type of problem can be solved with linear programming. 
When the learning problems on a data set are highly non-linear, 
each concept is learned independently with a complex non-linear classifier, for example, a deep neural network (DNN). The problem given in Eq.~(\ref{eq:general}) is best formulated as a multi-concept attack problem that can be modeled as a multi-objective optimization problem. 

Given a set of classification functions $f_{k}(x)$ where $k \in \{1, \ldots, n\}$ for $n$ concepts,  the multi-objective problem is formulated to find a perturbation for instance $x$ so that individual loss functions $\ell_1$ and $\ell_2$ are simultaneously maximized in a feasible region $\mathcal{X} \subset \mathbb{R}^n$: 
\begin{align*}
\max & \quad L(\Delta x) = (\ell_1(\Delta x), \ell_2(\Delta x))^T \\
%\max & \; \sum_{\forall k_i \in J}L(f_{k_i}(x+\Delta x),y_{k_i}) + \sum_{k_j \in [1,n]/J} L(f_{k_j}(x+\Delta x),-y_{k_j})\\
%s.t. & \quad ||c \odot \Delta x|| < \delta
s.t. & \quad x + \Delta x \in \mathcal{X} 
\end{align*}
where,
\begin{align*}
    \ell_1(\Delta x) & = \sum_{\forall k_i \in J}L(f_{k_i}(x+\Delta x),y_{k_i}) \\
    \ell_2(\Delta x) & = \sum_{k_j \in [1,n]/J} L(f_{k_j}(x+\Delta x),-y_{k_j})
\end{align*}
in which we attack a subset J of concepts by maximizing the classification loss in $\ell_1$, and protect the rest of the concepts by maximizing the ``reverse classification'' loss in $\ell_2$. 

\subsection{Optimization for Linear Classifiers}
In this section, we present the solution to the multi-concept attack problem in which learning problems are linear. We assume the learning models are linear classification models $f_{k_i}(x) = \mathbf{w}_{k_i} (x) + b_{k_i}$. An optimal modification to a given instance $x$ is computed, assuming $L_1$, $L_{\infty}$, and $L_2$ norms. 

\subsubsection{$L_1$ Norm Minimization}
For $p \ge 1$, $L_p$ norm is generally defined as:
\[
||x||_p=(|x_1|^p+|x_2|^p+\ldots+|x_m|^p)^{\frac{1}{p}}.
\]
where $x =(x_1, x_2, \ldots, x_m)$ is a point in $m$-dimensional feature space. Hence, $L_1$ norm is $||x||_1=|x_1|+|x_2|+\ldots+|x_m|$. 

$L_1$ norm minimization is also known as the least absolute values method. Our  multi-concept attack problem is defined as the following single objective constrained optimization problem:
\begin{align*}
\min_{i \in A} & \quad (|c_1 \Delta x_1| + \ldots + |c_i \Delta x_i|+ \ldots) \\
s.t. 
    %& \quad y \cdot (\mathbf{w}_{k_1, \ldots, k_{\ell}} x' + b )  \le 0\\
     & \quad \forall k_i \in J, \quad y_{k_i} \cdot [\mathbf{w}_{k_i} (x + \Delta x) + b_{k_i}]  \le 0\\
     & \quad \forall k_j \in K, \quad y_{k_j} \cdot [\mathbf{w}_{k_j} (x + \Delta x) + b_{k_j}]  > 0 
%&\quad y \cdot (\mathbf{w}_{K-\{k_1, \ldots, k_{\ell}\}} x' + b )  \ge 0
\end{align*}
where $A$ is the index set of the features that can be modified, $J$ and $K$ denote the attacked model set and the protected model set respectively. Modification to the $i^{th}$ feature is denoted as $\Delta x_i$ and the corresponding cost is denoted as $c_i$. Note that $y_{k_i}$ is the label for a given $x$ in the $k_i^{th}$ problem  space. The problem can be reduced to a linear program as follows:
\begin{align*}
\min & \quad \sum_{i \in A}t_i \\
s.t. & \quad \forall i \in A, \quad c_i \Delta x_i \le t_i\\
& \quad \forall i \in A, \quad -c_i \Delta x_i \le t_i\\
& \quad \forall i \in A, \quad t_i \ge 0 \\
& \quad \forall k_i \in J, \quad y_{k_i} \cdot [\mathbf{w}_{k_i} (x+\Delta x) + b_{k_i}]  \le 0\\
& \quad \forall k_j \in K, \quad y_{k_j} \cdot [\mathbf{w}_{k_j} (x+\Delta x) + b_{k_j}]  > 0 
%& \quad y \cdot (\mathbf{w}_{k_1, \ldots, k_{\ell}} x' + b )  \le 0\\
%&\quad y \cdot (\mathbf{w}_{k-\{k_1, \ldots, k_{\ell}\}} x' + b )  \ge 0
\end{align*}
Note that if the goal is to attack the positive instances only, simply relax the  attack constraint from: \[\forall k_i \in J, \quad y_{k_i} \cdot [\mathbf{w}_{k_i} (x+\Delta x) + b_{k_i}]  \le 0\] to \[\forall k_i \in J, \quad \mathbf{w}_{k_i} (x+\Delta x) + b_{k_i}  \le 0.\] This allows to only attack positive samples without forcing false positives.

\subsubsection{$L_\infty$ Norm}
For $L_\infty$ norm minimization, the multi-concept attack problem is formulated as follows:
\begin{align*}
\min \quad \max_{i \in A} & (|c_1 \Delta x_1|, \ldots, |c_i \Delta x_i|, \ldots) \\
s.t. 
    %& \quad y \cdot (\mathbf{w}_{k_1, \ldots, k_{\ell}} x' + b )  \le 0\\
     & \quad \forall k_i \in J, \quad y_{k_i} \cdot [\mathbf{w}_{k_i} (x + \Delta x) + b_{k_i}]  \le 0\\
     & \quad \forall k_j \in K, \quad y_{k_j} \cdot [\mathbf{w}_{k_j} (x + \Delta x) + b_{k_j}]  > 0 
%&\quad y \cdot (\mathbf{w}_{K-\{k_1, \ldots, k_{\ell}\}} x' + b )  \ge 0
\end{align*}
Similarly, we can reduced the above problem to a linear program. The original problem is equivalent to the following linear program:
\begin{align*}
\min & \quad t \\
s.t. & \quad \forall i \in A, \quad c_i \Delta x_i \le t\\
& \quad \forall i \in A, \quad -c_i \Delta x_i \le t\\
& \quad t \ge 0\\
& \quad \forall k_i \in J, \quad y_{k_i} \cdot [\mathbf{w}_{k_i} \cdot (x+\Delta x) + b_{k_i}]  \le 0\\
& \quad \forall k_j \in K, \quad y_{k_j} \cdot [\mathbf{w}_{k_j} \cdot (x+\Delta x) + b_{k_j}]  \ge 0 \\
%& \quad y \cdot (\mathbf{w}_{k_1, \ldots, k_{\ell}} x' + b )  \le 0\\
%&\quad y \cdot (\mathbf{w}_{k-\{k_1, \ldots, k_{\ell}\}} x' + b )  \ge 0
\end{align*}

\subsubsection{$L_2$ Norm}

For $L_2$ norm minimization, the original multi-concept attack problem is defined as follows:
\begin{align*}
\min_{i \in A} & \quad (|c_1 \Delta x_1|^2 + \ldots + |c_i \Delta x_i|^2+ \ldots) \\
s.t. 
    %& \quad y \cdot (\mathbf{w}_{k_1, \ldots, k_{\ell}} x' + b )  \le 0\\
     & \quad \forall k_i \in J, \quad y_{k_i} \cdot [\mathbf{w}_{k_i} (x + \Delta x) + b_{k_i}]  \le 0\\
     & \quad \forall k_j \in K, \quad y_{k_j} \cdot [\mathbf{w}_{k_j} (x + \Delta x) + b_{k_j}]  > 0 
%&\quad y \cdot (\mathbf{w}_{K-\{k_1, \ldots, k_{\ell}\}} x' + b )  \ge 0
\end{align*}
The problem can be cast as a dual problem with a Lagrangian as:
\begin{flalign*}
L(\Delta x, \lambda_1, \ldots, \lambda_K) & =  (c\Delta x)^T(c\Delta x) + \sum_{k_i\in J} \lambda_{k_i} y_{k_i} (\mathbf{w}_{k_i} (x+\Delta x) + b_{k_i}) \\
&  \quad  -\sum_{k_j \in K}\lambda_{k_j} y_{k_j}(\mathbf{w}_{k_j} (x+\Delta x) + b_{k_j})\\
& = (c\Delta x)^T(c\Delta x) + \sum_{k_i\in J} \lambda_{k_i} y_{k_i} \mathbf{w}_{k_i} \Delta x \\
& \quad -\sum_{k_j \in K}\lambda_{k_j} y_{k_j} \mathbf{w}_{k_j} \Delta x 
 + \sum_{k_i\in J} \lambda_{k_i} y_{k_i} (\mathbf{w}_{k_i} x + b_{k_i}) \\
& \quad-\sum_{k_j \in K}\lambda_{k_j} y_{k_j}(\mathbf{w}_{k_j} x + b_{k_j})\\
\end{flalign*}
Suppose $Q=c^Tc$ is positive definite, let $A = c \Delta x$,
\[G = \sum_{k_i\in J} \lambda_{k_i} y_{k_i} \mathbf{w}_{k_i}  -\sum_{k_j \in K}\lambda_{k_j} y_{k_j} \mathbf{w}_{k_j},\] and \[B = \sum_{k_i\in J} \lambda_{k_i} k_{k_i} (\mathbf{w}_{k_i} x + b_{k_i})  -\sum_{k_j \in K}\lambda_{k_j} y_{k_j} (\mathbf{w}_{k_j} x + b_{k_j}),\] the dual function is:
\[\inf_{\Delta x} L(\Delta x, \lambda_1, \ldots, \lambda_K) = \inf_{\Delta x} A^TA+G c^{-1} A + B\]
The minimizer of the above quadratic form of $A$ is:
\[A^*= -\frac{1}{2}(Gc^{-1})^T.\]
Therefore, the minimum of the quadratic form is:
\[A^{*T}A^* + Gc^{-1}A^*+B = -\frac{1}{4} Gc^{-1}(Gc^{-1})^T+B\]
Hence the dual problem is given by:
\begin{align*}
%max & \quad -\frac{1}{4} (\sum_{k_i\in S} \lambda_{k_i} y \mathbf{w}_{k_i}  -\sum_{k_j \in K/S}\lambda_{k_j} y \mathbf{w}_{k_j})c^{-1}(c^{-1})^T(\sum_{k_i\in S} \lambda_{k_i} y \mathbf{w}_{k_i}  -\sum_{k_j \in K/S}\lambda_{k_j} y \mathbf{w}_{k_j})^T+B \\
\max & \quad -\frac{1}{4} Gc^{-1}(c^{-1})^TG^T+B \\
s.t. & \quad \lambda_{k_i \in J, k_j \in K} \ge 0
\end{align*}
%-----------------------------------

\subsection{Optimization for Non-linear Classifiers}
\label{subsec:multi-obj}
In practice, we often encounter more complex problems in which the learning functions are highly non-linear. When the decision functions are non-linear, we can formulate the multi-concept attack problem as a multi-objective  optimization problem, where the objectives are potentially in conflict with one another. 

In the multi-concept attack problem, we are given a set of $n$ objectives including  maximization of classification losses for the attacked classifiers and minimization of classification losses for the protected classifiers, where $n$ is the total number of classifiers in consideration. Let the decision functions of the classifiers be $f_{i \in [1,n]}(\Delta x)$ and $t \in \mathbb{R}$ be the optimal value and $d \in \mathbb{R}^m$ be the common descent direction of the gradients if it exists, we can find such a common descent direction by minimizing the first-order Pareto stationarity~\cite{article_Fliege}:
\begin{align*}
\argmin_{d \in \mathbb{R}^m, t \in \mathbb{R}} 
& \quad t + \frac{1}{2}||d||^2\\
s.t. & \nabla f_i(\Delta x)^Td - t \le 0, \forall i \in [1,n]
\end{align*}

The dual of the above problem is~\cite{article_Liu}:
\begin{align*}
\argmin_{\lambda^n} \;& ||\sum_{i=1}^{n} \lambda_i \nabla f_i(\Delta x)||^2\\
s.t. \;& \lambda \in \Delta^n 
\end{align*}
where $\Delta^n =\{\lambda: \sum_{i=1}^{n} \lambda_i = 1, \forall i \in [1,n]\}$. For simplicity and efficiency, in this paper, we do not search for the optimal $\lambda$, but simply use a negative multi-gradient $g=-\sum_{i=1}^{n}\lambda_i \nabla f_i(\Delta x)$ with  preset $\lambda_i$ values.  

Our multi-objective optimization moves in the direction of common gradient descent $g$ guided by losses of both the attacked and protected classifiers. We achieve this goal by maximizing the summation of classification loss corresponding to the ground-truth label of all classifiers in the attacked classifier set and the classification loss corresponding to the negation of the ground-truth label of all the classifiers in the protected classifier set. 
Since we consider binary classifiers, the negative of the ground truth label in the case of the protected classifier set translates to the opposite label. We model this in the form of the following multi-objective formulation:
\[
\max \; \frac{1}{M} \sum_{\forall k_i \in M}L(f_{k_i}(x+\Delta x),y) + \frac{1}{N} \sum_{\forall k_j \in N} L(f_{k_j}(x+\Delta x),1-y)
\] In this formulation, we have $K$ independent binary classifiers and we attack $M$ out of them and preserve the remaining $N$ classifiers. The set of classification functions $f_{k_i}(x)$ belong to the attack set $M$ and $f_{k_j}(x)$ belongs to the protected set $N$. In the scenario where more than one classifier is in the attacked classifier set ($M$ > 1) or in the protected classifier set ($N$ > 1), we average the losses of all classifiers in that set. 

As seen from the formulation above, we are masking the labels of our protected classifiers while generating adversarial examples for our attacked classifiers. In \ref{sec:multi}, we show that by guiding the gradient direction $g$ appropriately, we can preserve the accuracy of underlying protected concepts without a significant drop in attack success of the attacked concept. As a result of label masking, in some instances, the accuracy of protected classifiers may increase compared to their original accuracy. 

\section{Experimental Results}
\label{sec:exp}

We demonstrate the effectiveness of our multi-concept attack techniques on several image datasets. We experimented on the MNIST dataset, selectively attacking a set of linear SVM classifiers with hinge loss.
%, fashion MNIST, and Face Recognition. 
In the non-linear case, we used two more complex datasets: Celeb and UTKFace, selectively attacking deep neural networks.

\subsection{Selective Attack against Linear Classifiers}
Given a set of linear SVM classifiers, we tested the following scenarios: 1.) attack one classifier while protecting the rest; 2.) attack more than one classifier while protecting the rest. The concepts being attacked and protected were randomly chosen. The experiments were performed on the MNIST dataset~\footnote{http://yann.lecun.com/exdb/mnist/}. 

We define three learning tasks on the MNIST data:
\begin{itemize}
\itemsep=0pt
\item[a.)] learning even/odd digits (i.e., concept EVEN);
\item[b.)] learning digits greater than or equal to 5 (i.e., concept $\ge 5$);
\item[c.)] learning zero/non-zero digits (i.e., concept ZERO).
\end{itemize}  
We essentially transform the original learning problem into a multi-concept problem with three sub-tasks. For example, digit ``0'' can be labeled as 'zero', 'even', and 'not >= 5' simultaneously. We randomly select 200 samples from the independent test set to attack/protect the randomly selected concepts.  

\subsubsection{Results of $L_1$ Norm Minimization}

\begin{figure}[!htb]
    \centering
    \begin{minipage}{0.245\textwidth}
        \includegraphics[trim={.1cm .5cm .1cm 0.5cm},clip,width=\textwidth]{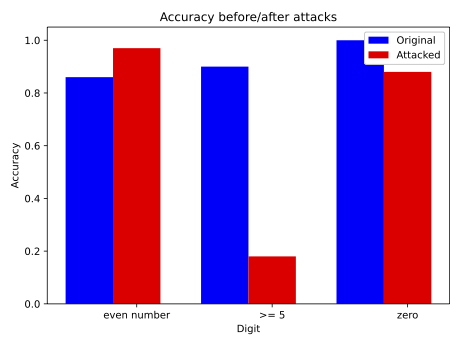}
%        \centering{\verb|Attack: zero|} \\
%        \centering{\verb|Protect: >=5, even number|}
    \end{minipage}%%
    \begin{minipage}{0.245\textwidth}
        \includegraphics[trim={0.1cm 0.5cm 0.1cm 0.5cm},clip,width=\textwidth]{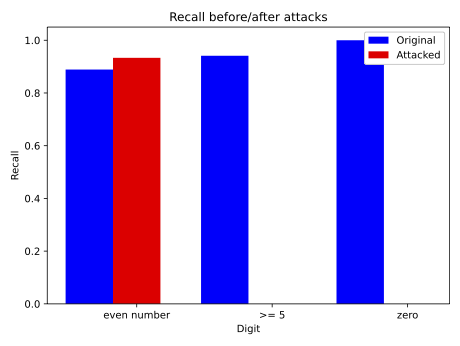}
%        \centering{\verb|Attack: even number, zero|} \\
%        \centering{\verb|Protect: >=5|}
    \end{minipage}
    \caption{\label{fig:mnist_l1_attk2} Multi-concept attack on the MNIST data with $L_1$ norm minimization. The concepts attacked are $\ge 5$ and {\em ZERO} and the protected concept is {\em EVEN}.}
\end{figure}

%*****************
% L1 attacked 2
%*****************
Figure~\ref{fig:mnist_l1_attk2} illustrates the results of the $L_1$ attack with two attack targets: concepts $\ge 5$ and {\em ZERO}. The concept {\em EVEN} is protected. The accuracy of the $\ge 5$ classifier dropped to less than 20\% from 88.5\%. The accuracy of the {\em ZERO} classifier dropped from 98\%  to less than 90\%, a significant drop considering there is only 10\% of digit `0' in the dataset. The recall values of both classifiers dropped to zero, while the accuracy/recall of the protected classifier slightly improved. These results confirmed our hypothesis that  multi-concepts can be attacked simultaneously while selectively protecting the others. 

\begin{figure*}[!htb]
    \begin{minipage}{0.333\textwidth}
       \includegraphics[trim={2.5cm 2.5cm 2.5cm 3cm},clip,width=\textwidth]{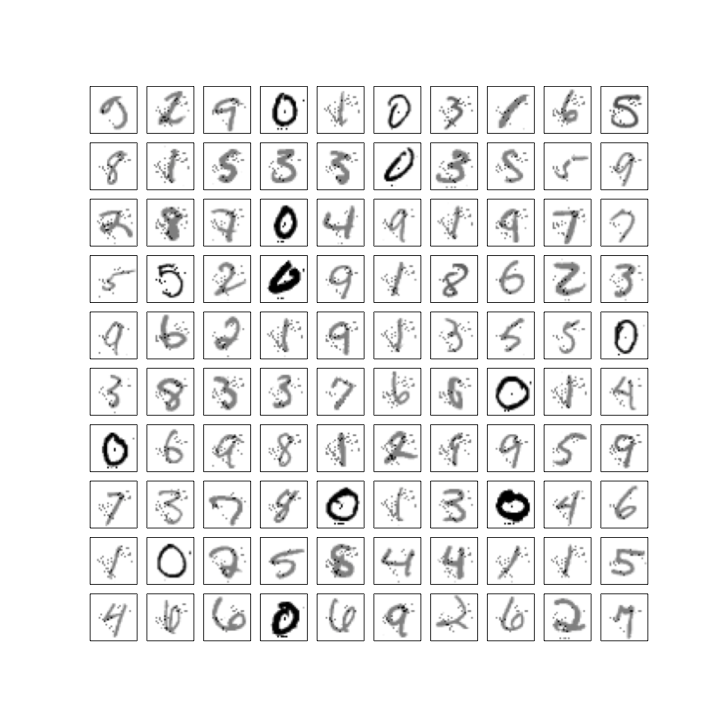}
        \centering{\verb|Attack: f(x)=">=5", f(x)="ZERO"|} \\
        %\caption{\able{fig:mnist_l1_attk2_imgs} Attack: $\ge 5$ and ZERO.}
%        \centering{\verb|Protect: >=5|}
    \end{minipage}
    \begin{minipage}{0.333\textwidth}
       \includegraphics[trim={5.5cm 2cm 5cm 2.5cm},clip,width=\textwidth]{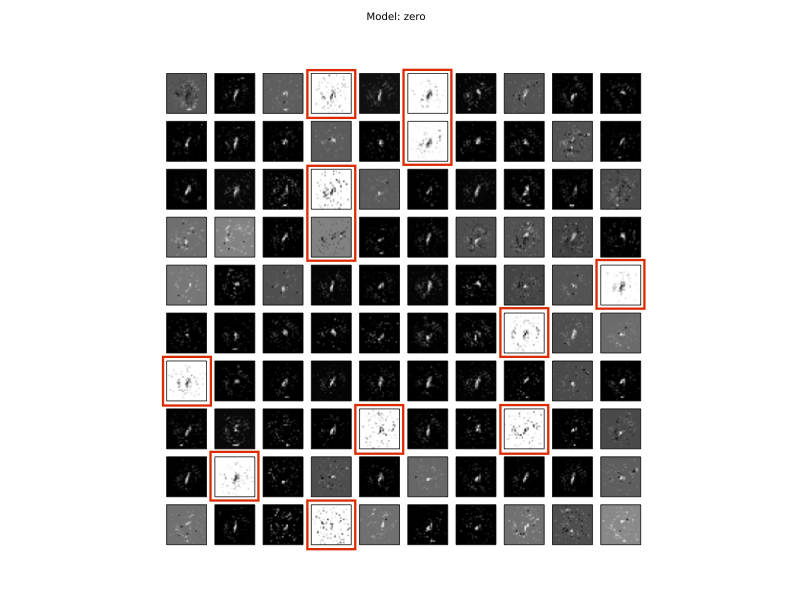}
        \centering{\verb|SHAP: f(x)="ZERO" on Original|} \\
%        \centering{\verb|Protect: >=5|}
    \end{minipage}
    \begin{minipage}{0.333\textwidth}
       \includegraphics[trim={5.5cm 2cm 5cm 2.5cm},clip,width=\textwidth]{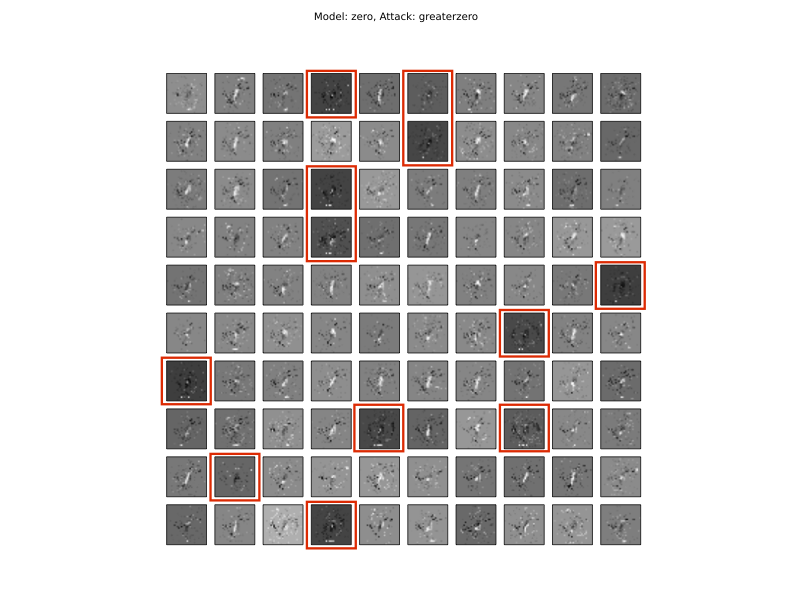}
    \centering{\verb|SHAP: f(x)="ZERO" on Attacked|} \\
%        \centering{\verb|Protect: >=5|}
    \end{minipage}
    \caption{\label{fig:mnist_l1_attk2_imgs} Multi-concept attack on the MNIST images with $L_1$ norm minimization with two concepts attacked. The left plot shows the attacked images, the middle plot shows the SHAP values of the pixels in the original images, and the right plot shows the SHAP values of pixels in the same images after the attack. The predictor  is the "ZERO" classifier.}
\end{figure*}

%Figure~\ref{fig:mnist_l1_attk2_imgs} shows the images of the selected digits after the attack.   
\begin{figure*}[!htb]
    \begin{minipage}{0.245\textwidth}
        \includegraphics[trim={2.5cm 2.5cm 2.5cm 3cm},clip,width=\textwidth]{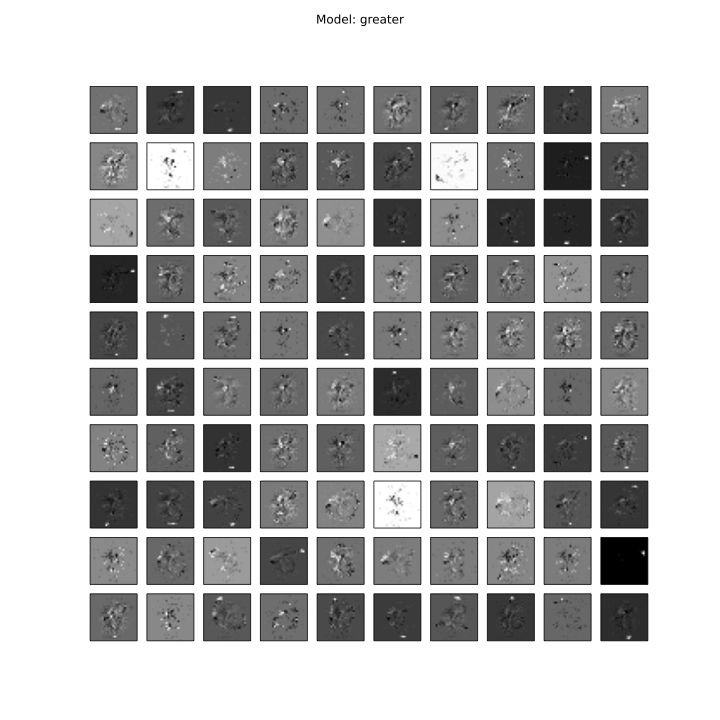}
        \centering{\verb|SHAP: f(x)=">=5"|}
        \centering{\verb|on Original|} 
    \end{minipage}%%
    \begin{minipage}{0.245\textwidth}
        \includegraphics[trim={2.5cm 2.5cm 2.5cm 3cm},clip,width=\textwidth]{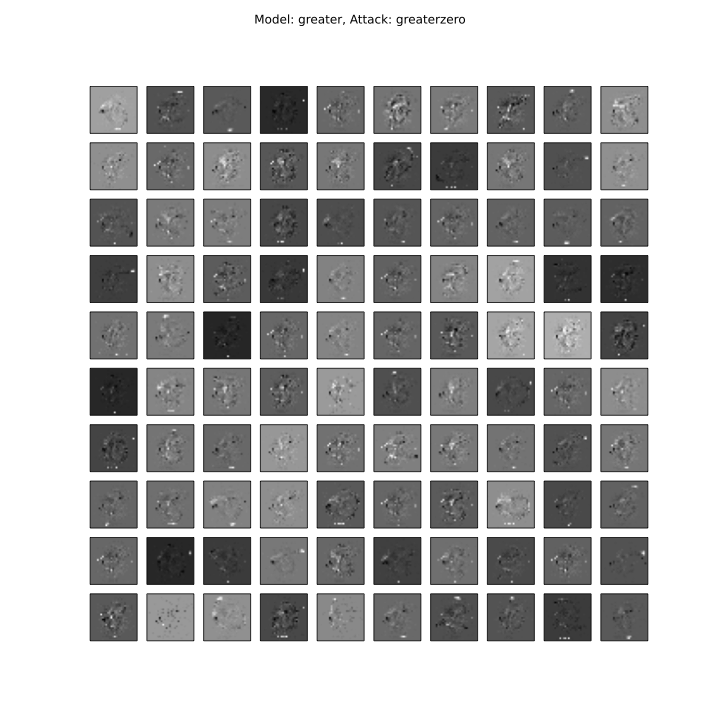}
        \centering{\verb|SHAP: f(x)=">=5"|}
        \centering{\verb|on Attacked|} 
    \end{minipage}%%
    \hfill
    %\centering{\verb|Attack: zero|} 
    %\centering{\verb|Protect: >=5, even number|}
    \begin{minipage}{0.245\textwidth}
        \includegraphics[trim={2.5cm 2.5cm 2.5cm 3cm},clip,width=\textwidth]{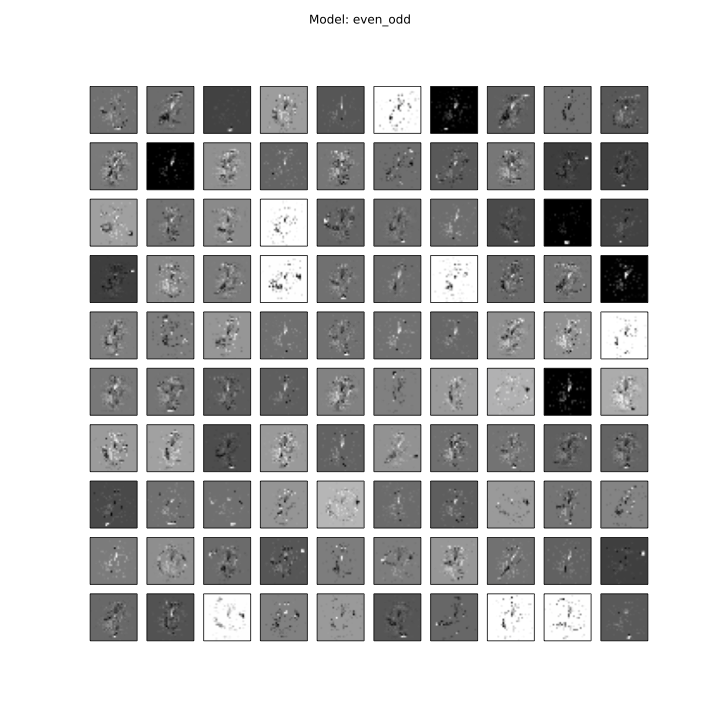}
        \centering{\verb|SHAP: f(x)="EVEN"|}
        \centering{\verb|on Original|} 
    \end{minipage}%%
    \begin{minipage}{0.245\textwidth}
        \includegraphics[trim={2.5cm 2.5cm 2.5cm 3cm},clip,width=\textwidth]{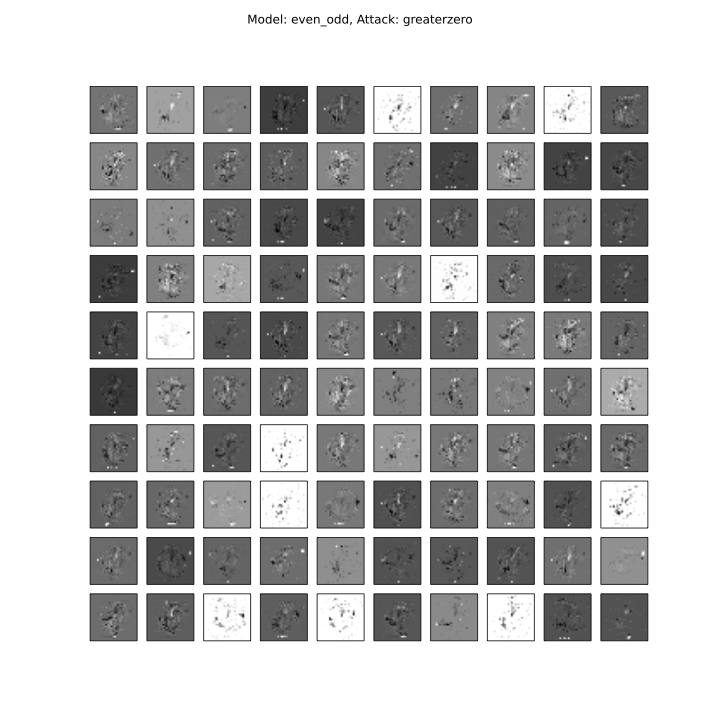}
        \centering{\verb|SHAP: f(x)="EVEN"|}
        \centering{\verb|on Attacked|} 
    \end{minipage}
    \caption{\label{fig:mnist_l1_attk2_protect} Multi-concept attack on the MNIST images with $L_1$ norm minimization. The left plot shows the SHAP values of the pixels in the original images and  in the attacked images with the classifier "$\ge 5$" (also attacked) as the predictor. The right plot shows the same SHAP values with the classifier "EVEN" (protected) as the predictor.}
\end{figure*}

We also investigate how the attacks influence the predictions of the  classifiers when  modifying the input images. We compute the Shapley values of each pixel in the original images and the images after the attack using SHAP~\cite{NIPS2017_7062}. SHAP (SHapley Additive exPlanations) is a game theoretic approach to explaining the decision of a machine learning model by providing the Shapley values from game theory. SHAP values represent each pixel's contribution to the decision output of the model. Small SHAP values indicate low contribution to the prediction of the concept.    Figure~\ref{fig:mnist_l1_attk2_imgs} shows the images after the attack, the SHAP values of the pixels in the original images, and the SHAP values in the images after the attack. We show the SHAP values of the positive class. Hence, pixels with larger SHAP values have significant contribution to the positive  prediction by the model, that is, a prediction of ZERO by the classifier trained to learn the {\em ZERO} concept. The smaller the SHAP values, the darker the pixels in the SHAP-value images, and hence the smaller contribution of the pixels  in the images of the digits to the prediction of the concept "ZERO". Likewise, the brighter the SHAP images, the more contribution of the pixels to the prediction of ``ZERO''.
    
From Figure~\ref{fig:mnist_l1_attk2_imgs}, we can clearly observe the following:
\begin{itemize}
\itemsep=0pt
    \item The SHAP values of the pixels in the original ``0'' images (red boxes) are much higher (brighter) than the pixels in images of other digits; while the SHAP values of the ``0'' pixels are much smaller (darker) in the attacked ``0'' images (red boxes). Also, notice the dark SHAP images (representing non-zero digits) on the original images (middle plot) turned out much lighter in the SHAP images (right plot) after the attack.   
    \item The attacked images of ``0'' are in fact made more perceptible to human eyes while compromising the classifier ZERO. This calls for more precautions in situations where human-AI cooperation is desired. Misunderstanding of AI behavior may result in upsetting consequences, especially in mission-critical applications such as self-driving vehicles.  
\end{itemize}

Figure~\ref{fig:mnist_l1_attk2_protect} shows the SHAP values of the image pixels when the predictions were made by the $\ge 5$ classifier and the EVEN classifier, respectively. Although not as obvious to human eyes as in Figure~\ref{fig:mnist_l1_attk2_imgs}, we can still see the colors of the SHAP values are toggled before and after the attacks. This explains the zero recall values shown in Figure~\ref{fig:mnist_l1_attk2}.

\begin{figure}[!htb]
    \centering
    \begin{minipage}{0.245\textwidth}
        \includegraphics[trim={.1cm .5cm .1cm 0.5cm},clip,width=\textwidth]{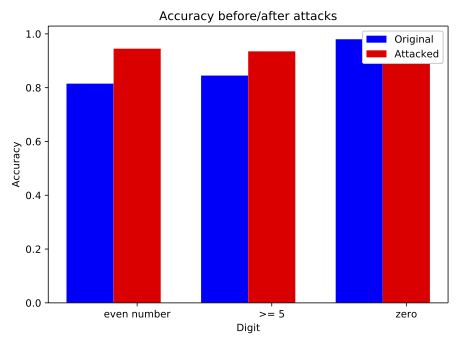}
%        \centering{\verb|Attack: zero|} \\
%        \centering{\verb|Protect: >=5, even number|}
    \end{minipage}%%
    \begin{minipage}{0.245\textwidth}
        \includegraphics[trim={0.1cm 0.5cm 0.1cm 0.5cm},clip,width=\textwidth]{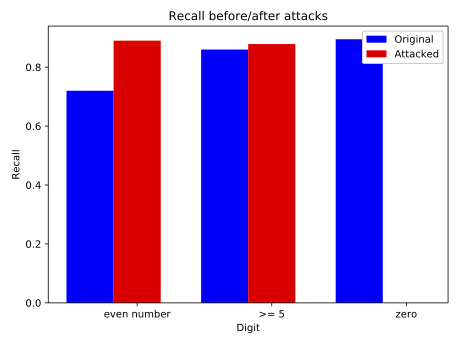}
%        \centering{\verb|Attack: even number, zero|} \\
%        \centering{\verb|Protect: >=5|}
    \end{minipage}
    \caption{\label{fig:mnist_l1_attk1} Multi-concept attack on the MNIST data with $L_1$ norm minimization. The concept attacked is {\em ZERO} and the protected concepts are {\em EVEN} and $\ge 5$.}
\end{figure}

\begin{figure}[!htb]
    \centering
    \begin{minipage}{0.245\textwidth}
        \includegraphics[trim={.1cm .5cm .1cm 0.5cm},clip,width=\textwidth]{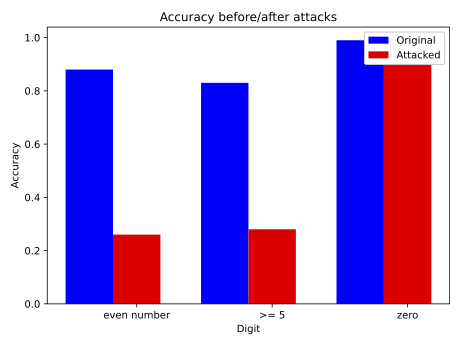}
%        \centering{\verb|Attack: zero|} \\
%        \centering{\verb|Protect: >=5, even number|}
    \end{minipage}%%
    \begin{minipage}{0.245\textwidth}
        \includegraphics[trim={0.1cm 0.5cm 0.1cm 0.5cm},clip,width=\textwidth]{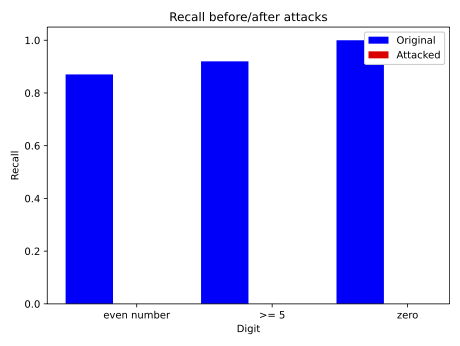}
%        \centering{\verb|Attack: even number, zero|} \\
%        \centering{\verb|Protect: >=5|}
    \end{minipage}
    \caption{\label{fig:mnist_l1_attk3} Multi-concept attack on the MNIST data with $L_1$ norm minimization. All three concepts {\em EVEN}, $\ge 5$, and {\em ZERO} are attacked.}
\end{figure}

When one concept (randomly chosen) is attacked, the resulting accuracy and recall are shown in Figure~\ref{fig:mnist_l1_attk1}, where the concept ZERO was attacked. Note that the data set contains only about 10\%  images of ``0'', hence a drop of accuracy from 98\% to 90\% is very significant, which is further confirmed by the zero recall values of ``0'' after the attack. The protected concepts remained intact, or became better in some cases. The results when all three concepts are attacked are shown in Figure~\ref{fig:mnist_l1_attk3}. All attacked classifiers were successfully deceived by the attacked images. The corresponding SHAP images are provided in the Appendix.  

\subsubsection{Results of $L_\infty$ Norm Minimization}

\begin{figure}[!htb]
    \centering
    \begin{minipage}{0.245\textwidth}
        \includegraphics[trim={.1cm .5cm .1cm 0.5cm},clip,width=\textwidth]{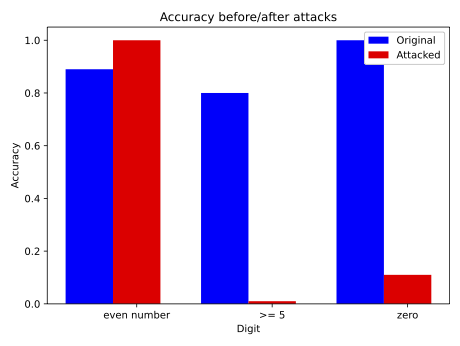}
%        \centering{\verb|Attack: zero|} \\
%        \centering{\verb|Protect: >=5, even number|}
        \end{minipage}%%
    \begin{minipage}{0.245\textwidth}
        \includegraphics[trim={0.1cm 0.5cm 0.1cm 0.5cm},clip,width=\textwidth]{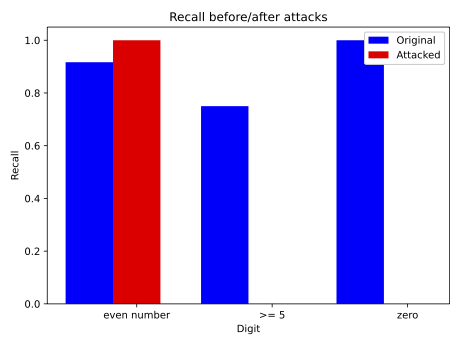}
%        \centering{\verb|Attack: even number, zero|} \\
%        \centering{\verb|Protect: >=5|}
        \end{minipage}
    \caption{\label{fig:mnist_linf_attk2} Multi-concept attack on the MNIST data with $L_\infty$ norm minimization. The concepts attacked are $\ge 5$ and {\em ZERO} and the protected concept is {\em EVEN}.}
\end{figure}

Figure~\ref{fig:mnist_linf_attk2} illustrates the results of the $L_\infty$ attack. The ``$\ge 5$''  classifier and the ``ZERO'' classifier were attacked, and the ``{\em EVEN}'' classifier was protected. As in the case of $L_1$ attack, the $L_{\infty}$ attacks improved the overall accuracy and recall of the protected classifier while slashing the accuracy and recall of the attacked classifier to nearly zero.

%********************************
\begin{comment}
    \begin{figure}[!htb]
        \centering
        \begin{minipage}{0.245\textwidth}
        \includegraphics[trim={.1cm .5cm .1cm 0.5cm},clip,width=\textwidth]{figures/new_eps/png/mnist_linf_attk1_acc.png}
%        \centering{\verb|Attack: zero|} \\
%        \centering{\verb|Protect: >=5, even number|}
        \end{minipage}%%
        \begin{minipage}{0.245\textwidth}
        \includegraphics[trim={0.1cm 0.5cm 0.1cm 0.5cm},clip,width=\textwidth]{figures/new_eps/png/mnist_linf_attk1_rec.png}
%        \centering{\verb|Attack: even number, zero|} \\
%        \centering{\verb|Protect: >=5|}
        \end{minipage}
        \caption{\label{fig:mnist_linf_attk1} Multi-concept attack on the MNIST data with $L_1$ norm minimization. The concepts attacked are $\ge 5$ and {\em ZERO} and the protected concept is {\em EVEN}.}
    \end{figure}
\end{comment}
%*********************************

\begin{figure}[!htb]
    \centering
    \includegraphics[trim={2.5cm 2.5cm 2.5cm 3cm},clip,width=0.225\textwidth]{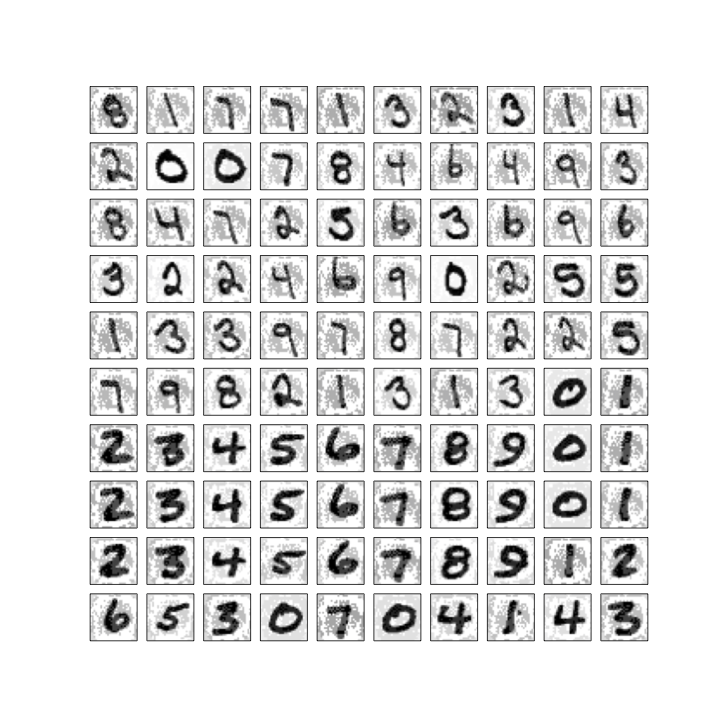}
    \caption{\label{fig:mnist_linf_attk2_imgs} MNIST images attacked with $L_\infty$ norm minimization. The concepts attacked are $\ge 5$ and {\em ZERO} and the protected concept is {\em EVEN}.}
\end{figure}

Figure~\ref{fig:mnist_linf_attk2_imgs} shows the images after the attack. To human eyes, the $L_{\infty}$ attacks are more aggressive than the $L_1$ attacks when modifying the images. The SHAP images (shown in the Appendix) were similar as in the case of the $L_1$ attacks. The results for attacking one and three concepts are also shown in the Appendix. 
   
\subsubsection{Results of $L_2$ Norm Minimization}

\begin{figure}[!htb]
    \centering
    \begin{minipage}{0.245\textwidth}
    \includegraphics[trim={.1cm .5cm .1cm 0.5cm},clip,width=\textwidth]{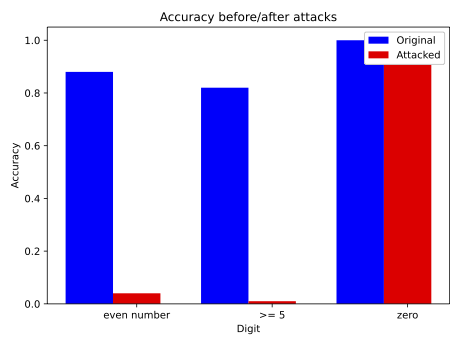}
%        \centering{\verb|Attack: zero|} \\
%        \centering{\verb|Protect: >=5, even number|}
        \end{minipage}%%
    \begin{minipage}{0.245\textwidth}
        \includegraphics[trim={0.1cm 0.5cm 0.1cm 0.5cm},clip,width=\textwidth]{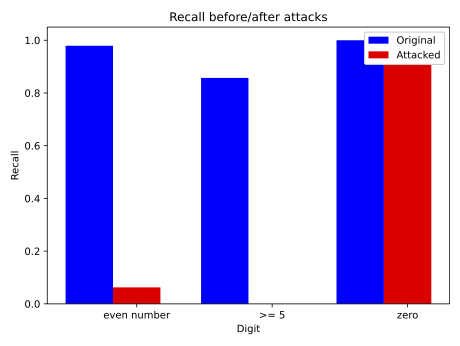}
%        \centering{\verb|Attack: even number, zero|} \\
%        \centering{\verb|Protect: >=5|}
        \end{minipage}
    \caption{\label{fig:mnist_l2_attk2} Multi-concept attack on the MNIST data with $L_2$ norm minimization. The concepts attacked are $\ge 5$ and {\em EVEN} and the protected concept is {\em ZERO}.}
\end{figure}

Figure~\ref{fig:mnist_l2_attk2} illustrates the results of the $L_2$ attack. The attacked concepts were {\em EVEN} and $\ge 5$, and the protected concept was {\em ZERO}. As in the studies of the $L_1$ and $L_{\infty}$ norm minimization, the $L_2$ attacks improved the overall accuracy of the protected classifier while tanking the performance of the classifiers under attack. 

\begin{figure}[!h]
    \centering
      \includegraphics[trim={2.5cm 2.5cm 2.5cm 3cm},clip,width=0.225\textwidth]{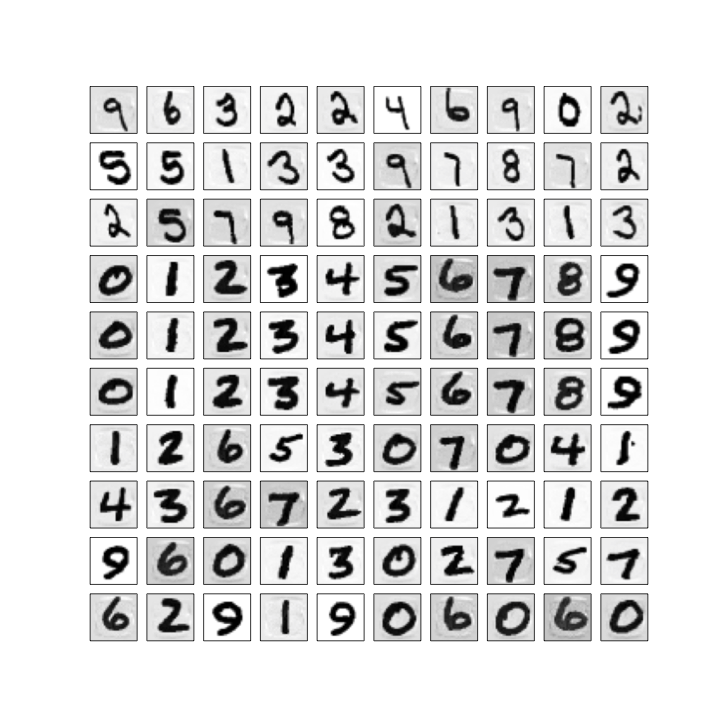}
    \centering{\verb|Attack: f(x)=">=5", f(x)="EVEN"|} \\
    \caption{\label{fig:mnist_l2_attk2_imgs} MNIST images attacked with $L_2$ norm minimization. The concepts attacked are {\em EVEN} and $\ge 5$ and the protected concept is {\em ZERO}.}
\end{figure}

As shown in Figure~\ref{fig:mnist_l2_attk2_imgs}, the $L_2$ attacks attempt to add noise evenly to the images than $L_1$ and $L_{\infty}$ attacks. The results for attacking one and three concepts are shown in the Appendix.    

\subsection{Selective Attack against Non-Linear Classifiers}
\label{sec:multi}

In this section, we selectively attack non-linear classifiers on more complicated image data. More specifically, we examine the effect on accuracy for:
\begin{itemize}
\itemsep=0pt
    \item[1.)] {\em one} attacked and {\em one}  protected classifier scenario
    \item[2.)] {\em one} attacked classifier and {\em two} protected classifiers scenario
    \item[3.)] {\em two} attacked classifiers and {\em one} protected classifier scenario.
\end{itemize}

We refer to these scenarios in short as $A:\{X\},\; P:\{Y\}$, where X is a set of {\em attacked} classifiers, Y is a set of {\em protected} classifiers. \footnote{We leave more complex scenarios of {\em multi}-attacked and {\em multi}-protected classifiers as future work.}. In the following sections, we show that our custom formulation not only minimizes the accuracy drop for protected classifiers, in some instances, it also increases the accuracy of the protected concepts in comparison to the baseline accuracy.

We choose CelebA~\cite{CelebA} and UTKFace~\cite{UTKFace} datasets for our experiments. For the CelebA dataset~\cite{CelebA}, we train three classifiers that predict the {\em GENDER}, {\em GLASSES}, and {\em AGE} concepts and use them to test our custom attacks. In the case of the UTKFace~\cite{UTKFace} dataset, we use classifiers that predict the {\em AGE}, {\em GENDER}, and {\em ETHNICITY} concepts. For the {\em AGE} concept, during the training process, we label all examples with {\em AGE} <= 30 as 0 and {\em AGE} > 30 as 1. For the {\em ETHNICITY} concept, we assign label 0 to class "White" and label 1 to the remaining classes. To train our classifiers, we use a pre-trained ResNet50~\cite{resnet} model architecture with a custom layer concatenating Average Pooling and Max Pooling layers, a Flatten layer, and two blocks of Batch normalization~\cite{BatchNorm}, Dropout~\cite{Dropout}, Linear and ReLU layers. We employ fastai~\cite{fastai} for training our models on the CelebA and UTKFace datasets. Table~\ref{tab:Training} reports the training results.

\begin{table}[!t]
\centering
\caption{Training and Validation accuracy for different classifiers on the CelebA and UTKFace dataset}
\begin{tabular}[t]{@{}llr@{}}
\toprule
& Training Accuracy & Validation Accuracy \\
\midrule
\textbf{CelebA} & & \\
Glasses & 99.71\% & 99.58\% \\
Gender & 97.33\% &  97.44\%  \\
Age & 88.68\% & 86.78\% \\
\midrule
\textbf{UTKFace} & & \\
Age & 93.44\% & 97.03\% \\
Gender & 99.49\% &  99.14\% \\
Ethnicity & 99.86\% & 99.88\% \\
\bottomrule
\label{tab:Training}
\end{tabular}
\end{table}

We compare our results with the following baselines: 1) Original accuracy, 2.) Adversarial accuracy after PGD~\cite{madry2019deep} $L_2$ attack and 3.) Adversarial accuracy after PGD $L_\infty$ attack. We use the Foolbox~\cite{foolbox} library to implement baseline PGD $L_2$ and PGD $L_\infty$ attacks.  After running for multiple epsilon values, we found that $\epsilon$ = 4 and $\epsilon$ = 0.3 produce the strongest PGD $L_2$ and $L_\infty$ attacks for our datasets. Hence, for our experiments, we use $\epsilon$ = 4 for $L_2$ norm attacks and $\epsilon$ = 0.3 for $L_\infty$ norm attacks. Since $L_\infty$ norm attacks are generally stronger attacks than $L_2$ norm attacks, a smaller $\epsilon$ value suffices for $L_\infty$ norm attacks. 

After experimenting with different step sizes and iterations for our experiments, we settled on step\_size = 0.8 for attacks of $L_2$ norm and step\_size = 0.06 for $L_\infty$ norm attacks. Choosing these right step sizes helps guide our attack quickly to the maximum attack success rate for the PGD $L_2$ and $L_\infty$ norm attacks.

For our experiments, we fix the number of iterations to 200 and the $\lambda_i$ (as in ~\cite{article_Fliege}) value to 1. We will examine the impact of $\lambda_i$ in future work. To generate adversarial examples, we choose 1000 examples from each dataset for comparison.

We show the accuracy of targeted PGD attacks for the CelebA and UTKFace datasets and their impacts on other classifiers in Tables~\ref{tab:PGD_L2_inf_Celeb} and ~\ref{tab:PGD_L2_Linf_UTKFace}. In Table~\ref{tab:PGD_L2_inf_Celeb}, we observe that a targeted PGD $L_2$ attack on the {\em GENDER} concept also reduces the accuracy of {\em AGE} and {\em GLASSES} concepts. The accuracy of {\em GLASSES} drops from 98.69\% original accuracy to 88.42\% and the accuracy of {\em AGE} drops from 85.7\% to 64.05\%. Similarly in Table~\ref{tab:PGD_L2_Linf_UTKFace} for UTKFace, a PGD $L_\infty$ attack on {\em AGE}, reduces the accuracy of {\em GENDER} from 99\% to 50.2\% and the {\em ETHNICITY} concept from 100\% to 58.7\%. Our goal is to generate adversarial examples attacking a set of concepts without causing a significant drop accuracy of our desired protected concept set. Using the custom loss formulation approach discussed in the previous section, we successfully retain the accuracy of the protected concepts. 

\begin{table}[!t]
\centering
\caption{Accuracy of {\em AGE}, {\em GENDER} and {\em GLASSES} concepts using adversarial examples from PGD $L_2$ and $L_\infty$ attacks on the {\em GENDER} concept and {\em GLASSES} concept for CelebA Dataset.}

\begin{tabular}[t]{@{}llll@{}}
\toprule
% &  \multicolumn{3}{c}{Classifiers} \\
%\midrule
& Age & Gender & Glasses\\
\midrule
Original Accuracy & & & \\
& 85.7\% & 93.76\% & 98.69\% \\
\midrule
PGD Attack Accuracy & & & \\
& & & \\
PGD $L_2$ Gender & 64.05\% & 3.73\% & 88.42\% \\
PGD $L_2$ Glasses & 64.25\% & 82.18\% & 3.63\% \\
& & & \\
PGD $L_\infty$ Gender & 40.79\% & 0\% & 50.45\% \\
PGD $L_\infty$ Glasses & 43.2\% & 51.86\% & 0.2\% \\
\bottomrule
\label{tab:PGD_L2_inf_Celeb}
\end{tabular}
\end{table}

\begin{table}[ht]
\centering
\caption{Accuracy of {\em AGE}, {\em GENDER} and {\em ETHNICITY} concepts using adversarial examples from PGD $L_2$ and $L_\infty$ attacks on {\em GENDER} concept and {\em AGE} concept for UTKFace Dataset}
\begin{tabular}[t]{@{}lllr@{}}
\toprule
% &  \multicolumn{3}{c}{Classifiers} \\
%\midrule
& Age & Gender & Ethnicity \\ 
\midrule
Original Accuracy & & & \\
& 97.2\% & 99\% & 100\% \\
\midrule
PGD Attack Accuracy & & & \\
& & & \\
PGD $L_2$ on Gender & 82.7\% & 33.8\% & 80\% \\
PGD $L_2$ on Age & 40.3\% & 80.4\% & 79.2\% \\
& & & \\
PGD $L_\infty$ on Gender & 50\% & 7\% & 58.5\% \\
PGD $L_\infty$ on Age & 30\% & 50.2\% & 58.7\% \\
\bottomrule
\label{tab:PGD_L2_Linf_UTKFace}
\end{tabular}
\end{table}

\begin{comment}
In the multi-concept classification setting where each image can be mapped to multiple concepts, using state-of-the-art attacks such as %like 
PGD $L_2$ and $L_\infty$ to attack a particular concept may degrade the accuracy of other concepts.
\end{comment}

\subsubsection{Results of $L_2$ Norm Minimization}
\label{subsec:l2}

\begin{table}[!t]
\centering
\caption{Accuracy on Celeb data for {\em AGE}, {\em GENDER} and {\em GLASSES} using multi-concept attack with $L_2$ norm minimization. A represents concepts being Attacked, P denotes concepts being Protected. The accuracy of Protected concepts is highlighted in bold.}
\begin{tabular}[t]{@{}lllr@{}}
\toprule
& Age & Gender & Glasses \\
\midrule
Original Accuracy & & & \\
& 85.7\% & 93.76\% & 98.69\% \\
\midrule
PGD Attack Accuracy & & & \\
& & & \\
PGD $L_2$ Gender & 64.05\% & 3.73\% & 88.42\% \\
PGD $L_2$ Glasses & 64.25\% & 82.18\% & 3.63\% \\
\midrule
Custom Attack Accuracy & & & \\
& & & \\
A:Gender, P:Glasses & 69.59\% & 15.31\% & \textbf{99.6}\% \\
A:Glasses, P:Gender & 67.67\% & \textbf{99.09}\% & 17.22\% \\
& & & \\
A:Gender, P:\{Glasses,Age\} & \textbf{91.74}\% & 17.32\% & \textbf{99.3}\% \\
A:Glasses, P:\{Gender,Age\} & \textbf{92.25}\% & \textbf{96.98}\% & 10.17\% \\
& & & \\
A:\{Gender,Glasses\}, P:Age & \textbf{97.28}\% & 47.33\% & 28.0\% \\
A:\{Glasses,Age\}, P:Gender & 39.68\% & \textbf{99.09}\% & 26.59\% \\
A:\{Gender,Age\}, P:Glasses & 29.31\% & 29.41\% & \textbf{99.9}\% \\
\bottomrule
\label{tab:Celeb_L2}
\end{tabular}
\end{table}

\begin{table}[ht]
\centering
\caption{Accuracy on Celeb data for {\em PRETTY}, {\em GENDER} and {\em GLASSES} using multi-concept attack with $L_2$ norm minimization. A represents concepts being Attacked, P denotes concepts being Protected. The accuracy of Protected concepts is highlighted in bold.}
\begin{tabular}[t]{@{}lllr@{}}
\toprule
& Pretty & Gender & Glasses \\
\midrule
Original Accuracy & & & \\
& 81.77\% & 93.76\% & 98.69\% \\
\midrule
PGD Attack Accuracy & & & \\
& & & \\
PGD $L_2$ Gender & 59.62\% & 3.73\% & 88.42\% \\
PGD $L_2$ Glasses & 65.46\% & 82.18\% & 3.63\% \\
\midrule
Custom Attack Accuracy & & & \\
& & & \\
A:Gender, P:Glasses & 67.07\% & 15.31\% & \textbf{99.6}\% \\
A:Glasses, P:Gender & 69.79\% & \textbf{99.09}\% & 17.22\% \\
& & & \\
A:Gender, P:\{Glasses,Pretty\} & \textbf{76.33}\% & 11.78\% & \textbf{99.19}\% \\
A:Glasses, P:\{Gender,Pretty\} & \textbf{80.46}\% & \textbf{97.78}\% & 8.36\% \\
& & & \\
A:\{Gender,Glasses\}, P:Pretty & \textbf{88.52}\% & 40.48\% & 21.85\% \\
A:\{Glasses,Pretty\}, P:Gender & 58.71\% & \textbf{99.6}\% & 28.2\% \\
A:\{Gender,Pretty\}, P:Glasses & 48.54\% & 28.1\% & \textbf{100}\% \\
\bottomrule
\label{tab:Celeb_L2_v2}
\end{tabular}
\end{table}

\begin{table}[ht]
\centering
\caption{Accuracy on UTKFace data for {\em AGE}, {\em GENDER} and {\em ETHNICITY} using multi-concept attack with $L_2$ norm minimization. A represents concepts being Attacked, P denotes concepts being Protected. The accuracy of Protected concepts is highlighted in bold.}
\begin{tabular}[t]{@{}lllr@{}}
\toprule
 & Age & Gender & Ethnicity \\
\midrule
Original Accuracy & & & \\
& 97.2\% & 99\% & 100\% \\
\midrule
PGD Attack Accuracy & & & \\
& & & \\
PGD $L_2$ Gender & 82.7\% & 33.8\% & 80\% \\
PGD $L_2$ Age & 40.3\% & 80.4\% & 79.2\% \\
\midrule
Custom Attack Accuracy & & & \\
& & & \\
A:Gender, P:Age & \textbf{95}\% & 47.6\% & 79.7\% \\
A:Age, P:Gender & 53.9\% & \textbf{96}\% & 79.6\% \\
& & & \\
A:Age, P:\{Gender,Ethnicity\} & 48.2\% & \textbf{93.9}\% & \textbf{95.3}\% \\
A:Gender, P:\{Age,Ethnicity\} & \textbf{94.4}\% & 44.2\% & \textbf{96}\% \\
& & & \\
A:\{Gender,Ethnicity\}, P:Age & \textbf{95.3}\% & 56.5\% & 44.1\% \\
A:\{Age,Ethnicity\}, P:Gender & 64.1\% & \textbf{94.8}\% & 46.4\% \\
A:\{Gender,Age\}, P:Ethnicity & 64.4\% & 56.9\% & \textbf{97.7}\% \\
\bottomrule

\begin{comment}
\begin{center}
\resizebox{\columnwidth}{!}{
\begin{tabular}{|c|c|c|c|}
 \hline
 Attack & {\em AGE} & {\em GENDER} & Ethnicity \\ 
 \hline 
Original accuracy & 97.2\% & 99\% & 100\% \\
& & & \\
PGD $L_2$ Gender & 82.7\% & 33.8\% & 80\% \\
PGD $L_2$Age & 40.3\% & 80.4\% & 79.2\% \\
& & & \\
A:Gender, P:Age & \textbf{95}\% & 47.6\% & 79.7\% \\
A:Age, P:Gender & 53.9\% & \textbf{96}\% & 79.6\% \\
& & & \\
A:Age, P:\{Gender,Ethnicity\} & 48.2\% & \textbf{93.9}\% & \textbf{95.3}\% \\
A:Gender, P:\{Age,Ethnicity\} & \textbf{94.4}\% & 44.2\% & \textbf{96}\% \\
& & & \\
A:\{Gender,Ethnicity\}, P:Age & \textbf{95.3}\% & 56.5\% & 44.1\% \\
A:\{Age,Ethnicity\}, P:Gender & 64.1\% & \textbf{94.8}\% & 46.4\% \\
A:\{Gender,Age\}, P:Ethnicity & 64.4\% & 56.9\% & \textbf{97.7}\% \\
 \hline 
\end{tabular}
}
\label{tab:UTKFace_L2}
\end{center}
\end{comment}
\label{tab:UTKFace_L2}
\end{tabular}
\end{table}

Tables~\ref{tab:Celeb_L2},~\ref{tab:Celeb_L2_v2}, and~\ref{tab:UTKFace_L2} illustrate the results for $L_2$ norm multi-concept attack on the Celeb~\cite{CelebA} and the UTKFace~\cite{UTKFace} datasets. They show the Original Accuracy of different concepts as well as their accuracy when attacked. While PGD $L_2$ attack drastically impacts the accuracy of the attacked concepts, as an unintended consequence, it also causes a drop in the accuracy of non-attacked models. 

As can be observed in Table~\ref{tab:Celeb_L2}, for the attack-1-protect-1 scenario (Attack:{\em GLASSES}, Protect:{\em GENDER}) our $L_2$ norm minimization attack protects the accuracy of the {\em GENDER} concept (99.09\%  after our custom attack versus 82.18\% after the PGD attack on {\em GLASSES} versus 93.76\% original accuracy). Similarly, in Table~\ref{tab:UTKFace_L2}, for the attack-1-protect-1 scenario (Attack:{\em GENDER}, Protect:{\em AGE}), the accuracy of {\em AGE} fared better when protected using our custom loss formulation compared to the PGD $L_2$ attack on the {\em GENDER} concept (95\% after our custom attack versus 82.7\% after the PGD Attack on {\em GENDER} versus 97.2\% original accuracy). Thus we see that there is no significant degradation in the attack accuracy of the concepts being attacked in our attack-1-protect-1 scenario compared to regular PGD attacks.  

We also observe that we retain the accuracy of the protected concepts in both the attack-1-protect-2 and the attack-2-protect-1 scenarios. In the case of the attack-1-protect-2 scenario (Attack:{\em GENDER}, Protect:{{\em GLASSES}, {\em AGE}}), our loss formulation protects {\em AGE} better than the corresponding accuracy after the PGD $L_2$ attack on other concepts ({\em AGE} has 91.74\% after our custom attack versus 64.05\% after the PGD attack on {\em GENDER} versus 85.7\% original accuracy). We see similar results in the attack-2-protect-1 scenario where the {\em AGE} concept is sufficiently protected (97.28\% after our custom attack versus 64.25\% after the PGD attack on {\em GLASSES} versus 85.7\% original accuracy). Table~\ref{tab:Celeb_L2_v2} also lists the outcome of the attack-m-protect-n scenario for CelebA dataset, where we consider {\em PRETTY}, {\em GENDER} and {\em GLASSES} concepts. We observe that the attack-1-protect-2 scenario successfully protects the {\em PRETTY} concept's accuracy (76.33\% after our custom attack versus 65.46\% after the PGD attack on GLASSES versus 81.77\% original accuracy). We get similar results in the attack-2-protect-1 scenario (PRETTY concept's accuracy is 88.52\% after our custom formulation versus 65.46\% after the PGD attack on Glasses and  81.77\% original accuracy). 

This observation is consistent for the  UTKFace dataset as seen in Table~\ref{tab:UTKFace_L2} where the accuracy of both the {\em AGE} and {\em ETHNICITY} concept are preserved. The accuracy for the {\em AGE} concept is 94.4\% after our custom attack. This is in contrast to its accuracy of 40.3\% after the PGD $L_2$ attack on {\em GENDER}. The original accuracy for the {\em AGE} concept is 97.2\%, thus being successfully protected by our custom attack. Similarly, the accuracy of {\em ETHNICITY} is 96\% after our custom attack versus 80\% after the PGD attack on {\em GENDER}, the original accuracy being 100\%. using our custom attack in the attack-1-protect-2 scenario (Attack:{\em GENDER}, Protect:{\em AGE}, {\em ETHNICITY}).

To show the impact of our attack on the important attributes of protected concepts, we calculate SHAP~\cite{NIPS2017_7062} values on the protected concept from Captum's~\cite{captum} GradientSHAP method. We plot these values as a heatmap using Captum's~\cite{captum} visualization tool against a color map of reverse Gray background. Brighter pixels in the plot indicate a higher SHAP score and more contribution in the image decision for the protected classifier. The GradientSHAP method approximates the SHAP~\cite{NIPS2017_7062} values by computing the expected values of gradients and multiplying it with the difference between the input and the baseline. The GradientSHAP method can also be considered as an approximation of Integrated Gradients~\cite{IntegratedGradients} as it computes the expected values of gradients against different baselines. 

\begin{figure}[!ht]
\centering
\includegraphics[trim={0.2cm 0.2cm 0.2cm 0.2cm},clip,width=0.4\textwidth]{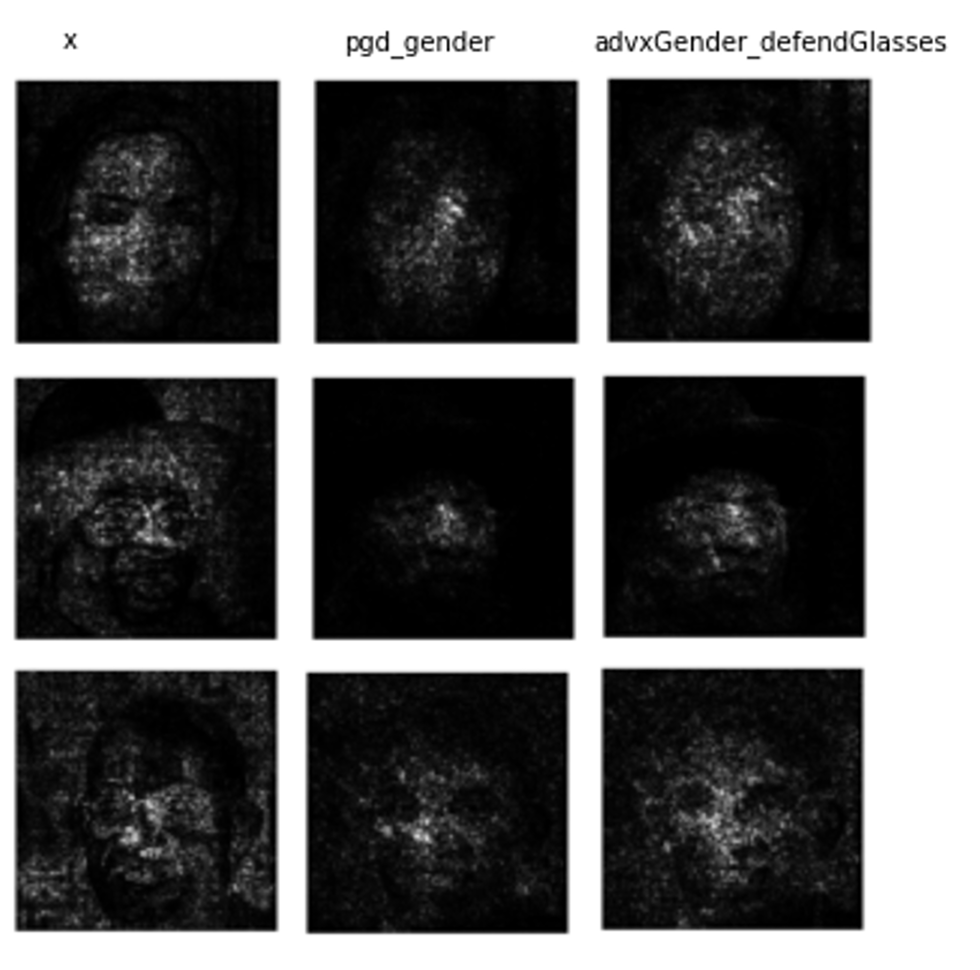}
\centering{\verb|Attack: f(x)=GENDER, Protect: f(x)=GLASSES|} \\
\caption{Visualizing SHAP values for CelebA images on {\em GLASSES} concept in three scenarios: ``x'', ``pgd\_gender'' and ``advxGender\_protectGlasses'' for $L_2$ norm minimization}
\label{fig:celeb_l2_gradShap_imgs} 
\end{figure}

Figure~\ref{fig:celeb_l2_gradShap_imgs} compares the calculated SHAP values as heat maps for the {\em GLASSES} concept and contrasts them between: a) The original normalized image, b) PGD $L_2$ attack image on {\em GENDER}, and c) Our custom attack image, attacking {\em GENDER} and protecting {\em GLASSES}. Case a) illustrates the important features that are picked up by the {\em GLASSES} model for prediction on the ground truth label. These features are not highlighted in a corresponding heat map image of case b) meaning that PGD $L_2$ adversarial attack on {\em GENDER} adds noise that also masks the important features for {\em GLASSES}. In c), our custom attack formulation successfully protects the said features and they are highlighted as important features by their brighter pixel values.

We also observe that increasing the number of attacked classifiers in our formulation decreases the protective strength of our attack on the concepts being defended. As shown in Sections~\ref{subsec:linf} and ~\ref{subsec:l2}, in Table~\ref{tab:Celeb_L2}, there's a slight drop in the {\em GENDER} concept accuracy ({\em GENDER} accuracy drops from 99.09\% versus 96.98\%) upon increasing the number of concepts being protected (only {\em GENDER} being protected versus protecting both {\em GENDER} and {\em AGE}). Despite having an accuracy drop, the accuracy of {\em GENDER} in our scenario is better than its accuracy after the PGD $L_2$ attack on {\em GLASSES} (96.98\% versus 82.18\%). 

\subsubsection{Results of $L_\infty$ Norm Minimization}
\label{subsec:linf}

\begin{table}[ht]
\centering
\caption{Accuracy on CelebA data for {\em AGE}, {\em GENDER} and {\em GLASSES} using multi-concept attack with $L_\infty$ norm minimization. In our custom attack, A represents concepts being Attacked, P denotes concepts being Protected. The accuracy of Protected concepts is highlighted in bold.}
\begin{tabular}[t]{@{}lllr@{}}
\toprule
 & Age & Gender & Glasses \\ 
\midrule
Original Accuracy & & & \\
& 85.7\% & 93.76\% & 98.69\% \\
\midrule
PGD Attack Accuracy & & & \\
& & & \\
PGD $L_\infty$ Gender & 40.79\% & 0\% & 50.45\% \\
PGD $L_\infty$ Glasses & 43.2\% & 51.86\% & 0.2\% \\
\midrule
Custom Attack Accuracy & & & \\
& & & \\
A:Gender, P:Glasses & 41.39\% & 0\% & \textbf{82.28}\% \\
A:Glasses, P:Gender & 58.61\% & \textbf{100}\% & 12.49\% \\
& & & \\
A:Gender, P:\{Glasses,Age\} & \textbf{98.79}\% & 0\% & \textbf{72.31}\% \\
A:Glasses, P:\{Gender,Age\} & \textbf{99.7}\% & \textbf{97.28}\% & 7.15\% \\
& & & \\
A:\{Gender,Glasses\}, P:Age & \textbf{100}\% & 10.88\% & 38.47\% \\
A:\{Glasses,Age\}, P:Gender & 2.32\% & \textbf{99.9}\% & 34.74\% \\
A:\{Gender,Age\}, P:Glasses & 0.91\% & 3.42\% & \textbf{88.52}\% \\
\bottomrule
\label{tab:Celeb_Linf}
\end{tabular}
\end{table}

Tables~\ref{tab:Celeb_Linf} and ~\ref{tab:UTKFace_Linf} show that in the case of $L_\infty$ norm minimization, our custom attack maintains the attack strength on the Attacked concepts while protecting the accuracy of the Protected concepts. For example, in Table~\ref{tab:UTKFace_Linf} for the attack-1-protect-1 attack scenario of A:{\em GENDER}, P:{\em AGE} (Attack Gender, Defend Age), our custom attack's attack strength is comparable to PGD $L_\infty$ attack on {\em GENDER} (7.5\% versus 7\%). It also successfully preserves the accuracy of the {\em AGE} concept compared to its corresponding accuracy for PGD $L_\infty$ attack on {\em AGE} (80.9\% versus 50\%). We also observe this in Table~\ref{tab:Celeb_Linf} in the scenario A:{\em GENDER}, P:{\em GLASSES} (Attacking the Gender concept and Protecting the Glasses concept). The accuracy of {\em GLASSES} is 82.28\% for our custom attack scenario versus 50.45\% on adversarial examples generated with the PGD $L_\infty$ attack on {\em GENDER}. This is achieved without a significant drop in the attack strength on the {\em GENDER} concept. (The {\em GENDER} accuracy dropped to 41.39\% using our custom attack versus 40.79\% for PGD $L_\infty$ attack on {\em GENDER}).

We also see similar results for other multi-concept sets of attacked/protected classifiers. For the 1-attack-2-protect scenario of A:{\em GLASSES}, P:{{\em GENDER}, {\em AGE}} from Table~\ref{tab:Celeb_Linf}, we successfully protect {\em GENDER} and {\em AGE} as a part of the Protected Set. Their accuracy after the custom attack is better than the original accuracy (97.28\% versus 93.76\% in case of the {\em GENDER} concept and 99.7\% versus 85.7\% for the {\em AGE} concept). At the same time, this increase in defense doesn't come at the cost of the attack strength for the {\em GLASSES} concept. It doesn't experience a significant degradation compared to the regular PGD $L_\infty$ attack on {\em GLASSES} (7\% versus 0\%). This observation is consistent for the {\em PRETTY} concept as shown in Table~\ref{tab:Celeb_Linf_v2}. In the attack-1-protect-2 scenario of A:{\em GENDER}, P:{{\em GLASSES}, {\em PRETTY}}, the Protected set consists of {\em PRETTY} and {\em GLASSES} while {\em GENDER} is the attacked concept. Here, not only is the {\em PRETTY} concept successfully protected without a significant drop in accuracy (79.05\% versus 81.77\% original accuracy), the attack strength of {\em GENDER} does not experience any drop for targeted PGD $L_\infty$ attack on {\em GENDER}.

\begin{table}[ht]
\centering
\caption{Accuracy on UTKFace data for {\em AGE}, {\em GENDER} and {\em ETHNICITY} using multi-concept attack with $L_\infty$ norm minimization. A represents concepts being Attacked, P denotes concepts being protected. The accuracy of protected concepts is highlighted in bold.}
\begin{tabular}[t]{@{}lllr@{}}
\toprule
 & Age & Gender & Ethnicity \\ 
\midrule
Original Accuracy & & & \\
& 97.2\% & 99\% & 100\% \\
\midrule
PGD Attack Accuracy & & & \\
& & & \\
PGD $L_\infty$ Gender & 50\% & 7\% & 58.5\% \\
PGD $L_\infty$ Age & 30\% & 50.2\% & 58.7\% \\
\midrule
Custom Attack Accuracy & & & \\
& & & \\
A:Gender, P:Age & \textbf{80.9}\% & 7.5\% & 57\% \\
A:Age, P:Gender & 23.4\% & \textbf{95.9}\% & 57.2\% \\
& & & \\
A:Age, P:\{Gender,Ethnicity\} & 21.1\% & \textbf{97.4}\% & \textbf{99.9}\% \\
A:Gender, P:\{Age,Ethnicity\} & \textbf{61.4}\% & 6.5\% & \textbf{99.6}\% \\
& & & \\
A:\{Age,Ethnicity\}, P:Gender & 45.8\% & \textbf{95.4}\% & 1.2\% \\
A:\{Gender,Age\}, P:Ethnicity & 43.9\% & 4.3\% & \textbf{100}\% \\
A:\{Gender,Ethnicity\}, P:Age & \textbf{83.7}\% & 3.6\% & 1.7\% \\
\bottomrule
\label{tab:UTKFace_Linf}
\end{tabular}
\end{table}

\begin{table}[ht]
\centering
\caption{Accuracy on CelebA data for {\em PRETTY}, {\em GENDER} and {\em GLASSES} using multi-concept attack with $L_\infty$ norm minimization. In our custom attack, A represents concepts being Attacked, P denotes concepts being protected. The accuracy of protected concepts is highlighted in bold.}
\begin{tabular}[t]{@{}lllr@{}}
\toprule
 & Pretty & Gender & Glasses \\ 
\midrule
Original Accuracy & & & \\
& 81.77\% & 93.76\% & 98.69\% \\
\midrule
PGD Attack Accuracy & & & \\
& & & \\
PGD $L_\infty$ Gender & 35.65\% & 0\% & 50.45\% \\
PGD $L_\infty$ Glasses & 68.08\% & 51.86\% & 0.2\% \\
\midrule
Custom Attack Accuracy & & & \\
& & & \\
A:Gender, P:Glasses & 52.06\% & 0\% & \textbf{82.28}\% \\
A:Glasses, P:Gender & 69.49\% & \textbf{100}\% & 12.49\% \\
& & & \\
A:Gender, P:\{Glasses,Pretty\} & \textbf{79.05}\% & 0\% & \textbf{71.9}\% \\
A:Glasses, P:\{Gender,Pretty\} & \textbf{89.53}\% & \textbf{99.7}\% & 6.14\% \\
& & & \\
A:\{Gender,Glasses\}, P:Pretty & \textbf{98.59}\% & 1.61\% & 42.3\% \\
A:\{Glasses,Pretty\}, P:Gender & 37.06\% & \textbf{100}\% & 39.98\% \\
A:\{Gender,Pretty\}, P:Glasses & 30.21\% & 0.0\% & \textbf{93.25}\% \\
\bottomrule
\label{tab:Celeb_Linf_v2}
\end{tabular}
\end{table}

\begin{figure}[!t]
\centering
\includegraphics[trim={0.2cm 0.2cm 0.2cm 0.2cm},clip,width=0.4\textwidth]{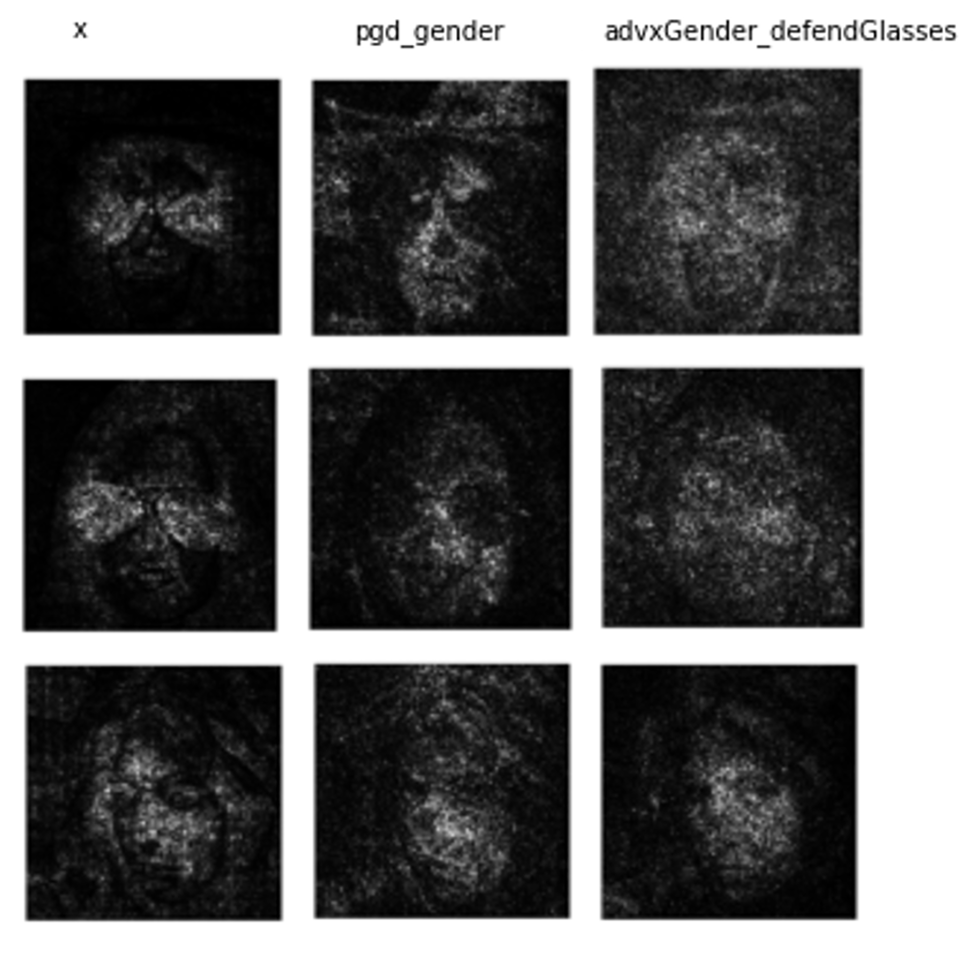}
\centering{\verb|Attack: f(x)=GENDER, Protect: f(x)=GLASSES|} 
\caption{\label{fig:celeb_linf_gradShap_imgs} Visualizing SHAP values for CelebA images on {\em GLASSES} concept in three scenarios: ``x'', ``pgd\_gender'' and ``advxGender\_protectGlasses'' for $L_\infty$ norm minimization}
\end{figure}

\begin{figure}[!ht]
\centering
\includegraphics[trim={0.2cm 0.2cm 0.2cm 0.2cm},clip,width=0.4\textwidth]{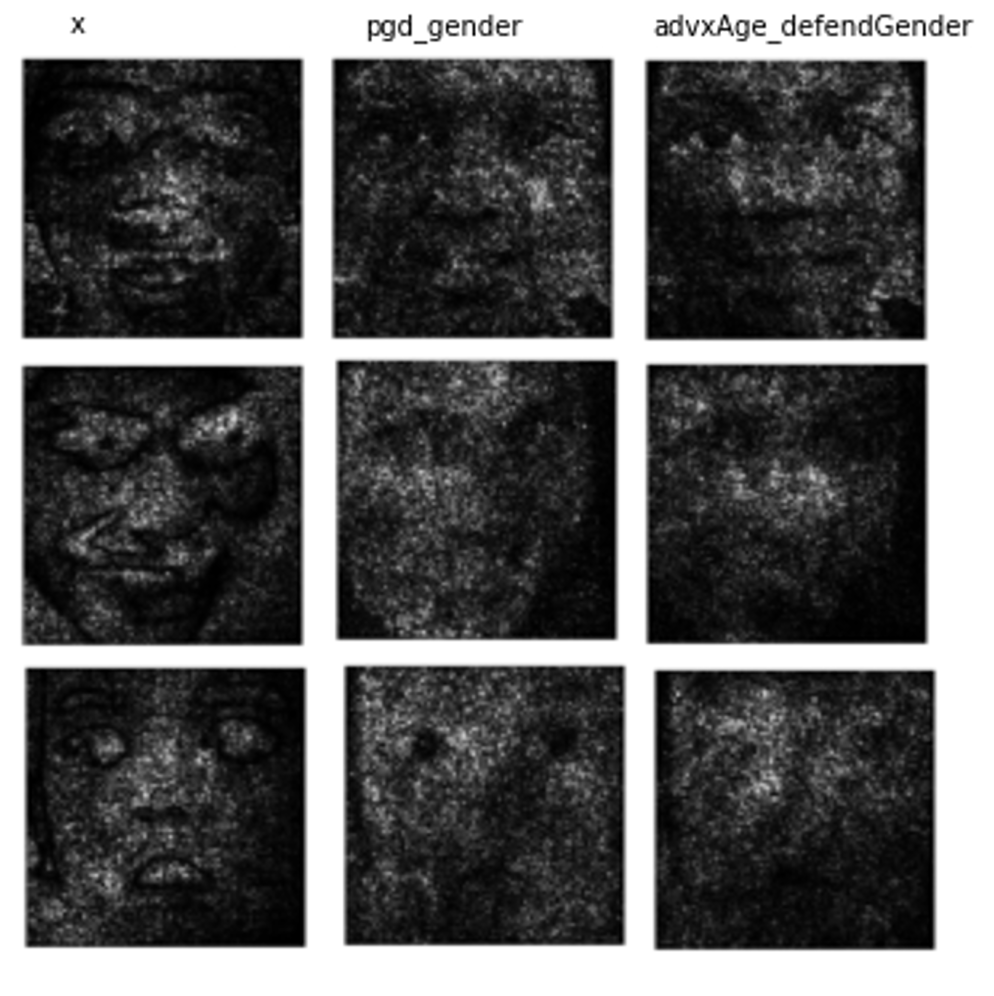}
\centering{\verb|Attack: f(x)=AGE, Protect: f(x)=GENDER|} 
\caption{\label{fig:utkFace_linf_gradShap_imgs} Visualizing SHAP values for UTKFace images on {\em GENDER} concept in three scenarios: ``x'', ``pgd\_Age'' and ``advxAge\_protectGender'' for $L_\infty$ norm minimization}
\end{figure}

We calculate SHAP values on images for the $L_\infty$ norm minimization and display the visualization in  Figure~\ref{fig:celeb_linf_gradShap_imgs} for CelebA dataset and Figure~\ref{fig:utkFace_linf_gradShap_imgs} for UTKFace dataset. From Figure~\ref{fig:celeb_linf_gradShap_imgs}, we see that adversarial examples generated using our custom attack successfully preserves features that are important for identifying the {\em GLASSES} concept. This is in contrast to PGD $L_\infty$ attack on {\em GENDER} where the corresponding bright pixels on the {\em GLASSES} concept are darkened. We also see that the SHAP values for $L_\infty$ norm minimization are sharper and more distinct compared to the $L_2$ norm minimization. This is because the $L_\infty$ norm minimization of the attack is stronger than the $L_2$ norm variant. In the case of the UTKFace dataset, PGD $L_\infty$ attack on {\em GENDER} adds a lot of noise throughout the image to generate the adversarial example and does not return a meaningful attribution. In contrast, our custom attack generates more targeted noise around important pixels for the {\em GENDER} concept.

Usually, attacks with the $L_\infty$ norm minimization are stronger than attacks with the $L_2$ norm minimization. The PGD $L_\infty$ attack accuracy for the {\em AGE} concept in Table~\ref{tab:UTKFace_Linf} is lower than the PGD $L_2$ attack accuracy (40.3\% vs 30\%) in Table~\ref{tab:UTKFace_L2}, thus signifying a higher attack success. We observe that this increase in attack strength comes at the expense of the accuracy of other non-attacked concepts. In the previously discussed scenario comparing the accuracy of {\em GENDER} concept, the accuracy of non-attacked concept of {\em GENDER} is 80.4\% for PGD $L_2$ attack on {\em AGE} and 50.2\% for PGD $L_\infty$ attack. However, for our custom attack-1-protect-1 scenario where the {\em AGE} concept is attacked and the {\em GENDER} concept is protected, there is not much drop in the accuracy of the protected classifiers for both $L_2$ norm and $L_\infty$ norm settings (96\% versus 95.9\%). Thus, the advantage of the $L_\infty$ norm minimization for the defense of the protected concepts is more pronounced compared to the $L_2$ norm minimization of our attack. This can be seen by contrasting the improvement in accuracy of the {\em GENDER} concept for $L_\infty$ norm (50.2\% versus 95.9\%) with the accuracy increase for the corresponding $L_2$ norm (80.4\% versus 96\%). 
%-------------------------------------------------------------------------------

%-------------------------------------------------------------------------------

\section{Conclusion}
\label{sec:conclusion}
In this paper, we present  multi-concept attack models targeting a set of classifiers. The goal of the multi-concept attack problem is to attack one set of classifiers and protect the rest without resulting in a significant drop in their accuracy. We motivate our work by showing that when an instance is assessed by multiple classifiers, attacking the classifier trained to learn one concept (e.g. {\em GENDER}) may reduce the accuracy of the classifier trained on another concept (e.g. {\em GLASSES}), either because of a potential correlation between multiple concepts or simply because of the added noise misleading other classifiers. To provide a theoretical formulation of our multi-concept attack problem, we use integer linear programming to solve scenarios with linear classifiers to generate adversarial examples at test time. We extend this to the multi-concept classification with non-linear classifiers and include the loss functions of both the attacked classifiers and the protected classifiers in our formulation. We present our experimental results on datasets where each instance belongs to different categories of concepts. We show that our approaches are successful in attacking targeted concepts while protecting others in both settings compared to the baseline attacks. In some cases, we observe that the protected classifier's accuracy is higher than its original accuracy. We also show that our attack strategy can successfully attack and protect classifiers regardless of the size of each of the concept sets. 
%-------------------------------------------------------------------------------

%-------------------------------------------------------------------------------
%\bibliographystyle{plain}
%\bibliography{\jobname}
\bibliographystyle{plain}
\bibliography{multi_attack}

\appendix

\section{SHAP values for Linear Classifiers}
%Figure~\ref{fig:mnist_l1_attk1} illustrates the results of the $L_1$ norm minimization  when only one concept is attacked. The attacked concept is EVEN and the protected concepts are ZERO and $\ge 5$. The left plot shows the accuracy of the ``EVEN'' classifier dropped significantly while the other two were not affected, and the right plot shows the recall of the ``EVEN'' classifier dropped  nearly to zero. 

\begin{figure*}[!htb]
    \begin{minipage}{0.333\textwidth}
        \includegraphics[trim={2.5cm 2.5cm 2.5cm 3cm},clip, width=\textwidth]{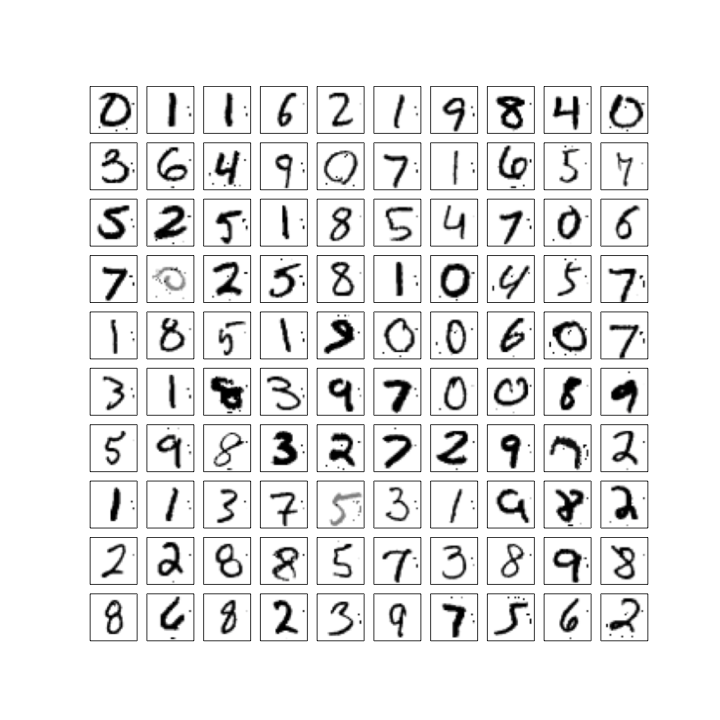}
        \centering{\verb|Attack: EVEN|} 
        \end{minipage}
    \begin{minipage}{0.333\textwidth}
        \includegraphics[trim={2.5cm 2.5cm 2.5cm 3cm},clip,width=\textwidth]{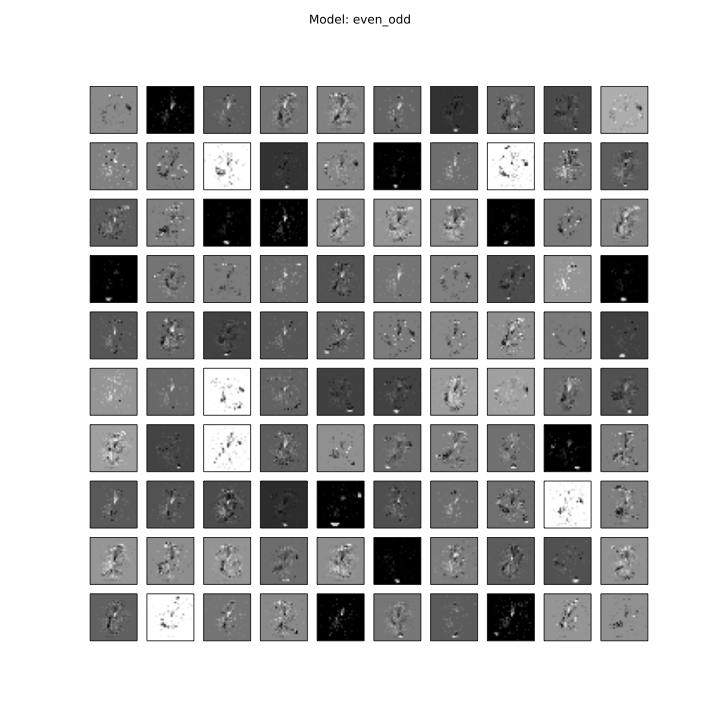}
        \centering{\verb|Exp. EVEN on Original|} 
        \end{minipage}
    \begin{minipage}{0.333\textwidth}
        \includegraphics[trim={2.5cm 2.5cm 2.5cm 3cm},clip,width=\textwidth]{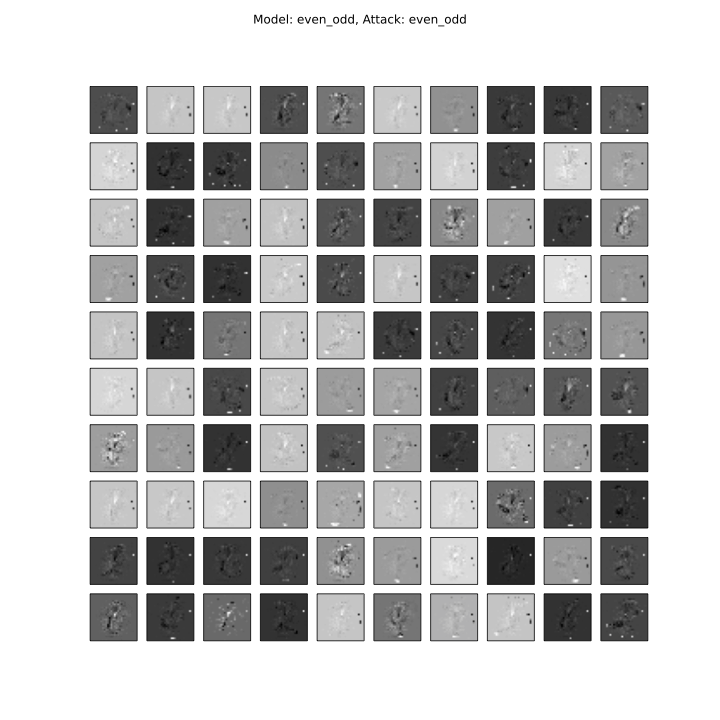}
        \centering{\verb|Exp. EVEN on Attacked|} 
        \end{minipage}
        %\centering{\verb|Attack: zero|} 
        %\centering{\verb|Protect: >=5, even number|}
    \caption{\label{fig:mnist_l1_attk1_imgs} Multi-concept attack on the MNIST images with $L_1$ norm minimization. The left plot shows the attacked images, the middle plot shows the SHAP values of the pixels in the original images, and the right plot shows the SHAP values of pixels in the same images after the attack. The predictor  is the "EVEN" classifier.}
\end{figure*}

Figure~\ref{fig:mnist_l1_attk1_imgs} illustrates the results of the $L_1$ norm minimization  when only one concept is attacked. The attacked concept is EVEN and the protected concepts are ZERO and $\ge 5$. The plots show the images of the selected digits after the attack. Although all the images are tainted with a noticeable amount of noise, images of `even' digits clearly stand out with darker appearances.   
    
\begin{figure*}[!htb]
    \begin{minipage}{0.245\textwidth}
        \includegraphics[trim={2.5cm 2.5cm 2.5cm 3cm},clip,width=\textwidth]{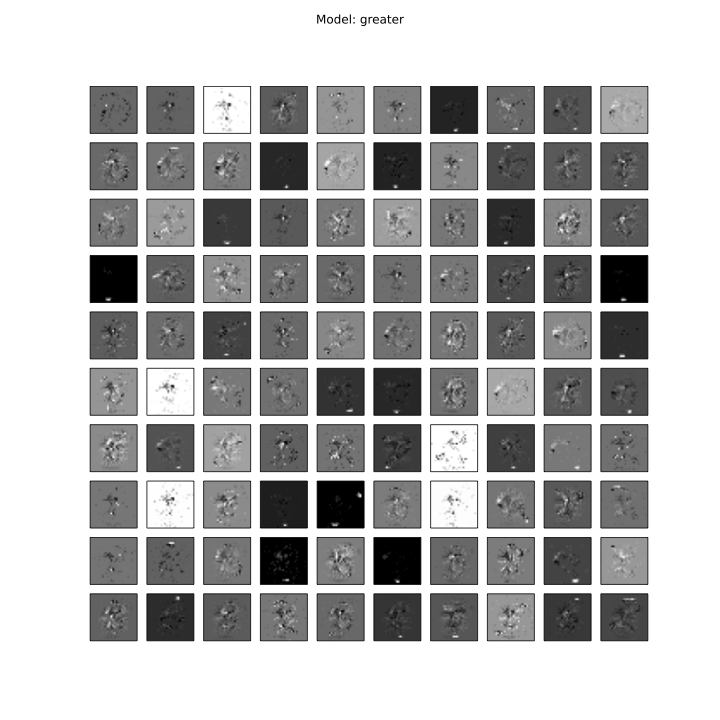}
        \centering{\verb|Exp. >=5 on Original|} 
    \end{minipage}
    \begin{minipage}{0.245\textwidth}
        \includegraphics[trim={2.5cm 2.5cm 2.5cm 3cm},clip,width=\textwidth]{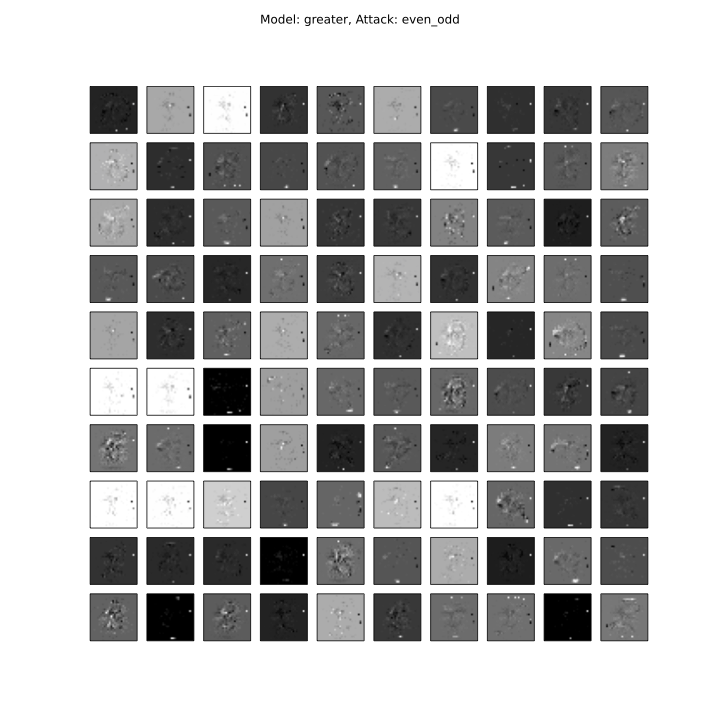}
        \centering{\verb|Exp. >=5 on Attacked|} 
    \end{minipage}
    \hfill
    %\centering{\verb|Attack: zero|} 
    %\centering{\verb|Protect: >=5, even number|}
    \begin{minipage}{0.245\textwidth}
        \includegraphics[trim={2.5cm 2.5cm 2.5cm 3cm},clip,width=\textwidth]{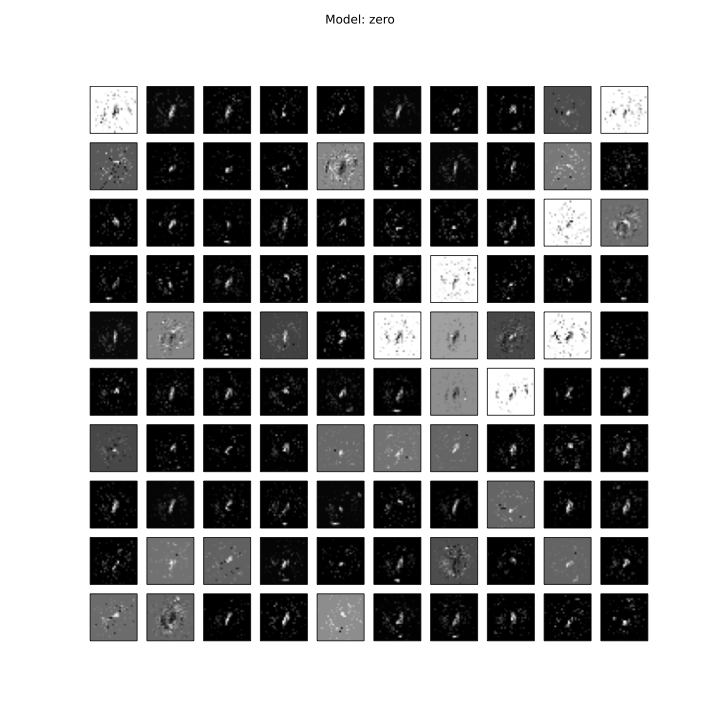}
        \centering{\verb|Exp. ZERO on  Original|} 
        \end{minipage}
    \begin{minipage}{0.245\textwidth}
        \includegraphics[trim={2.5cm 2.5cm 2.5cm 3cm},clip,width=\textwidth]{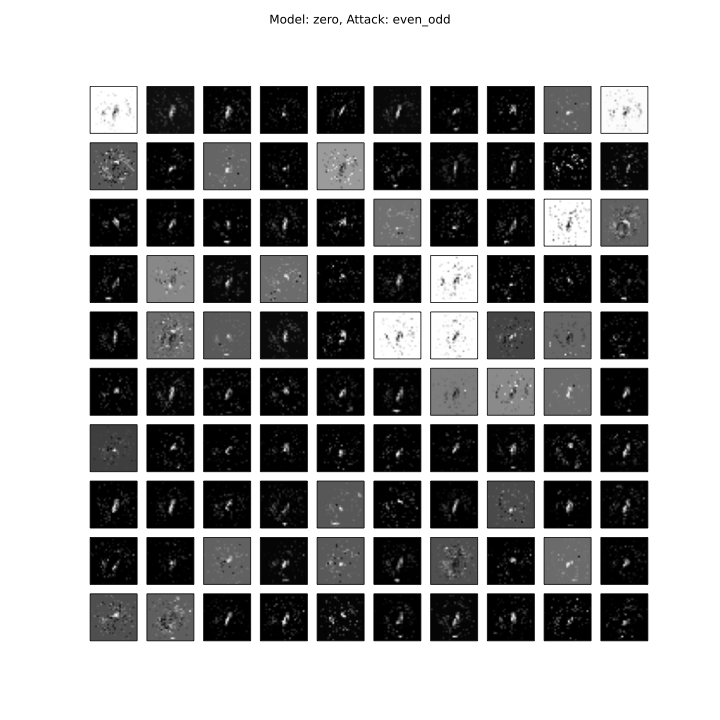}
        \centering{\verb|Exp. ZERO on  Attacked|} 
        \end{minipage}
    \caption{\label{fig:mnist_l1_attk1_protected} Multi-concept attack on the MNIST images with $L_1$ norm minimization. The left plot shows the SHAP values of the pixels in the original images and in the attacked images with the classifier "$\ge 5$" (protected) as the predictor. The right plot shows the same SHAP values with the classifier ``ZERO'' (also protected) as the predictor.}
\end{figure*}

Figure~\ref{fig:mnist_l1_attk1_protected} shows the Shapley values of the ``$\ge 5$'' and the ``ZERO'' classifiers, before and after the attack against the ``EVEN'' classifier. The Shapley values do not have significant changes as the two classifiers are protected. 

\begin{figure*}[!htb]
    \begin{minipage}{0.333\textwidth}
       \includegraphics[trim={2.5cm 2.5cm 2.5cm 3cm},clip,width=\textwidth]{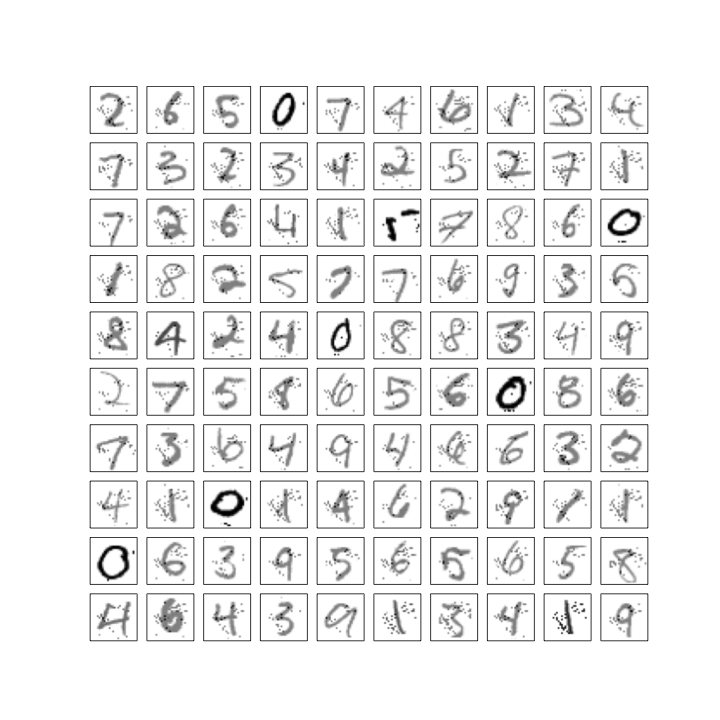}
        \centering{\verb|Attack: EVEN, >=5, ZERO|} \\
%        \centering{\verb|Protect: >=5|}
    \end{minipage}
    \begin{minipage}{0.333\textwidth}
       \includegraphics[trim={2.5cm 2.5cm 2.5cm 3cm},clip,width=\textwidth]{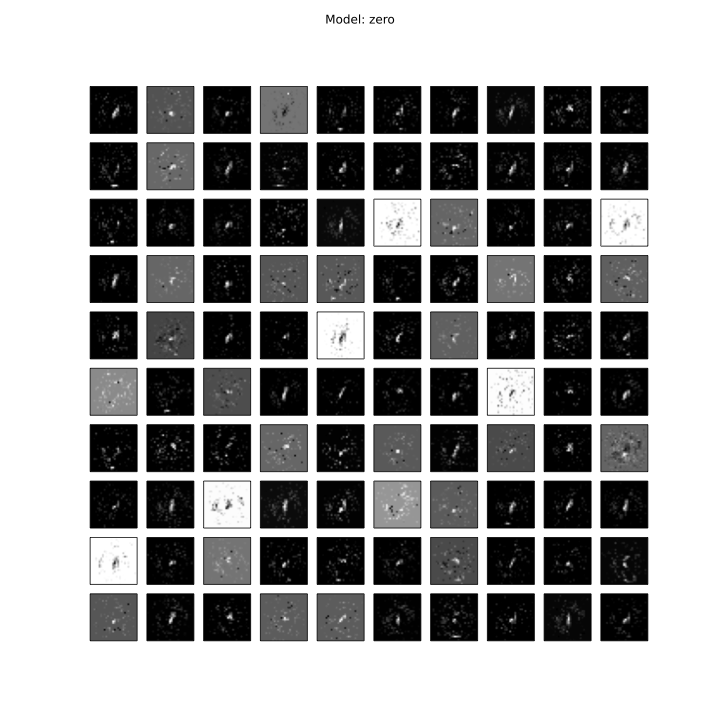}
        \centering{\verb|Exp. ZERO on Original|} \\
%        \centering{\verb|Protect: >=5|}
    \end{minipage}
    \begin{minipage}{0.333\textwidth}
       \includegraphics[trim={2.5cm 2.5cm 2.5cm 3cm},clip,width=\textwidth]{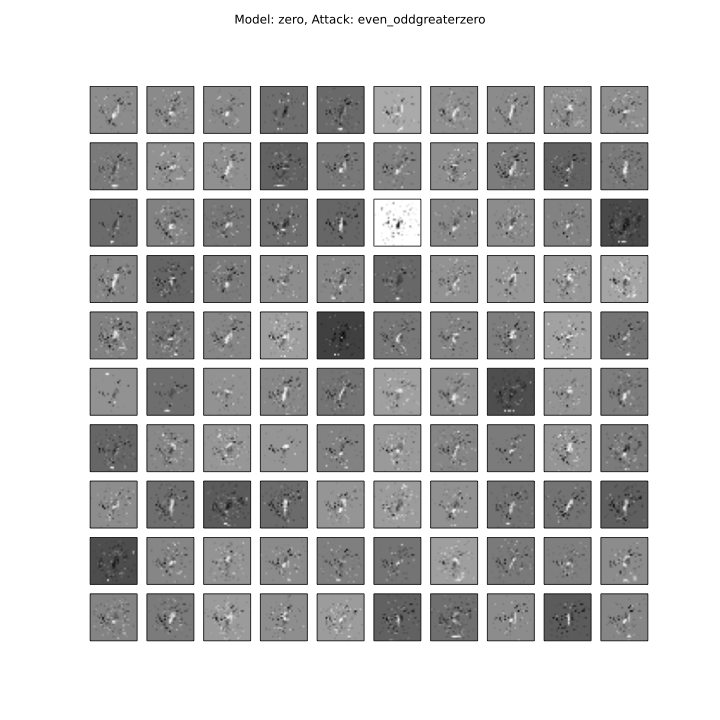}
        \centering{\verb|Exp. ZERO on Attacked|} \\
%        \centering{\verb|Protect: >=5|}
    \end{minipage}
    \caption{\label{fig:mnist_l1_attk3_imgs} Multi-concept attack on the MNIST images with $L_1$ norm minimization. The left plot shows the attacked images, the middle plot shows the SHAP values of the pixels in the original images, and the right plot shows the SHAP values of pixels in the same images after the attack. The predictor  is the "ZERO" classifier.}
\end{figure*}

\begin{figure*}[!htb]
    \begin{minipage}{0.245\textwidth}
        \includegraphics[trim={2.5cm 2.5cm 2.5cm 3cm},clip,width=\textwidth]{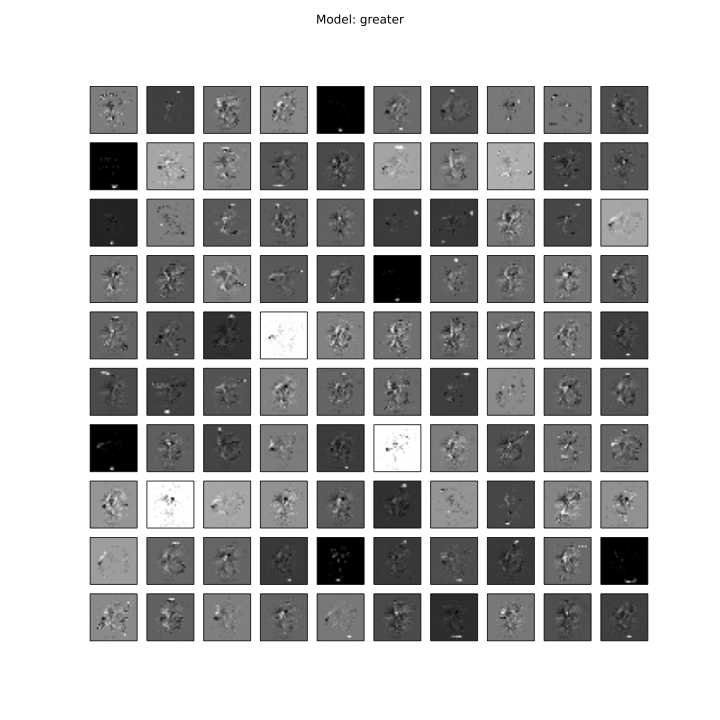}
        \centering{\verb|Exp. >=5 on Original|} 
        \end{minipage}%%
    \begin{minipage}{0.245\textwidth}
        \includegraphics[trim={2.5cm 2.5cm 2.5cm 3cm},clip,width=\textwidth]{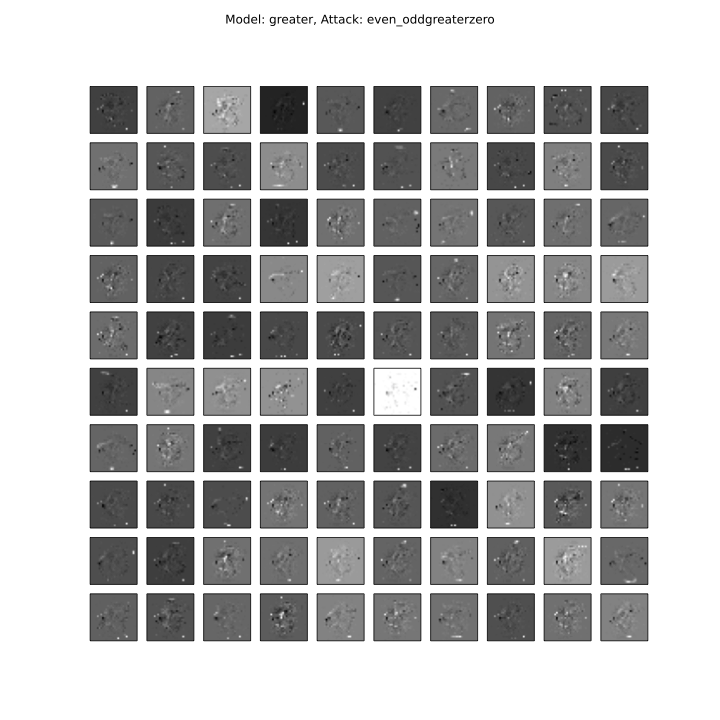}
        \centering{\verb|Exp. >=5 on Attacked|} 
        \end{minipage}%%
    \hfill
    %\centering{\verb|Attack: zero|} 
    %\centering{\verb|Protect: >=5, even number|}
    \begin{minipage}{0.245\textwidth}
        \includegraphics[trim={2.5cm 2.5cm 2.5cm 3cm},clip,width=\textwidth]{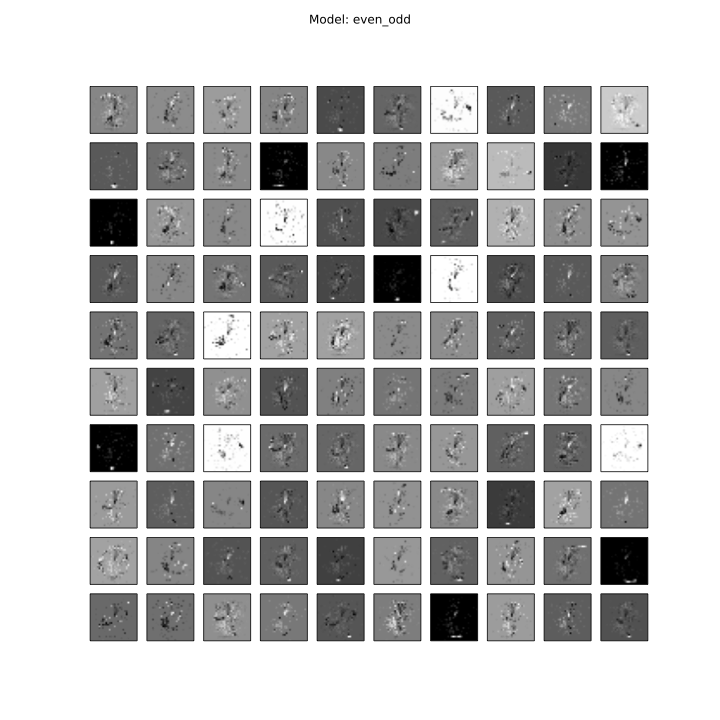}
        \centering{\verb|Exp. EVEN on  Original|} 
        \end{minipage}%%
    \begin{minipage}{0.245\textwidth}
        \includegraphics[trim={2.5cm 2.5cm 2.5cm 3cm},clip,width=\textwidth]{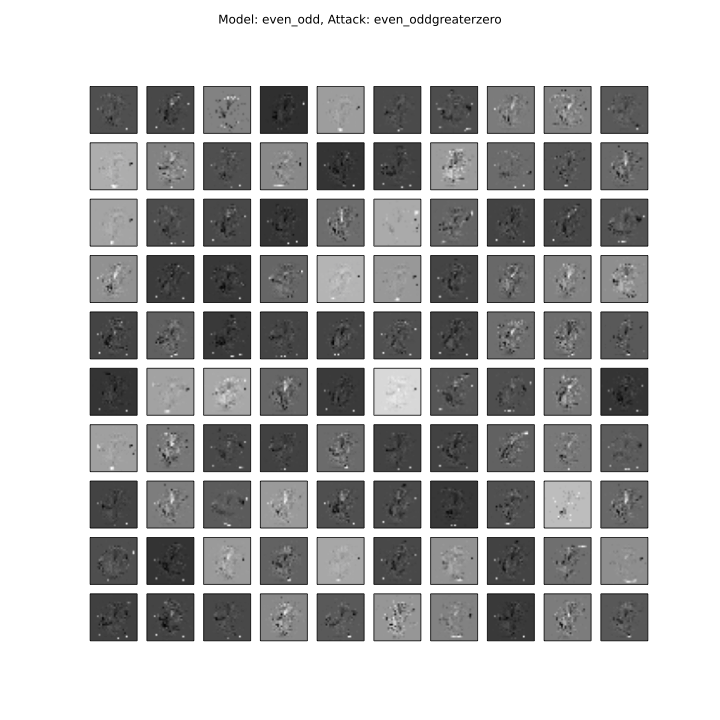}
        \centering{\verb|Exp. EVEN on  Attacked|} 
        \end{minipage}
    \caption{\label{fig:mnist_l1_attk3_protected} Multi-concept attack on the MNIST images with $L_1$ norm minimization. The left plot shows the SHAP values of the pixels in the original images and in the attacked images with the classifier "$\ge 5$" (attacked) as the predictor. The right plot shows the same SHAP values with the classifier ``EVEN'' (also attacked) as the predictor.}
\end{figure*}

When all three classifiers are attacked, as shown in Figures~\ref{fig:mnist_l1_attk3_imgs}-~\ref{fig:mnist_l1_attk3_protected}, it is clear the Shapley values have been toggled after the attack for all three classifiers.

\begin{figure}[!htb]
        \centering
        \begin{minipage}{0.245\textwidth}
        \includegraphics[width=\textwidth]{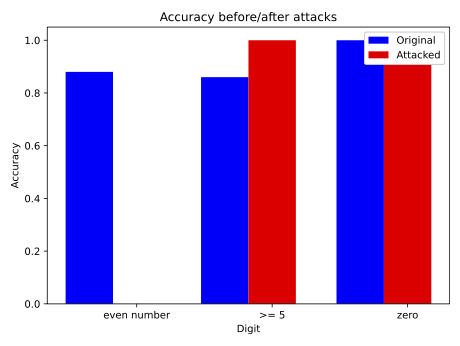}
%        \centering{\verb|Attack: zero|} \\
%        \centering{\verb|Protect: >=5, even number|}
        \end{minipage}%%
        \begin{minipage}{0.245\textwidth}
        \includegraphics[width=\textwidth]{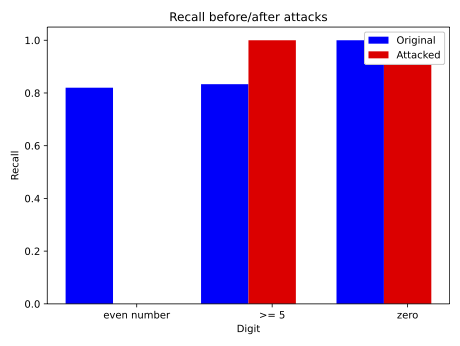}
%        \centering{\verb|Attack: even number, zero|} \\
%        \centering{\verb|Protect: >=5|}
        \end{minipage}
        \caption{\label{fig:mnist_l2_attk1} Multi-concept attack on the MNIST data with $L_2$ norm minimization. The concept attacked is {\em even} and the protected concepts are {\em zero} and {\em greater than/equal to 5}.}
\end{figure}

%*****************
% L2 attacked 1
%*****************
Figure~\ref{fig:mnist_l2_attk1} shows the accuracy and recall of the $L_2$ norm minimization when only one concept is attacked. The attacked concept is ``EVEN'' and the protected concepts are ``ZERO'' and ``>=5''. Both accuracy and recall of the EVEN classifier dropped significantly, while the other two classifiers are well protected. 

\begin{figure*}[!h]
    \begin{minipage}{0.333\textwidth}
       \includegraphics[trim={2.5cm 2.5cm 2.5cm 3cm},clip,width=\textwidth]{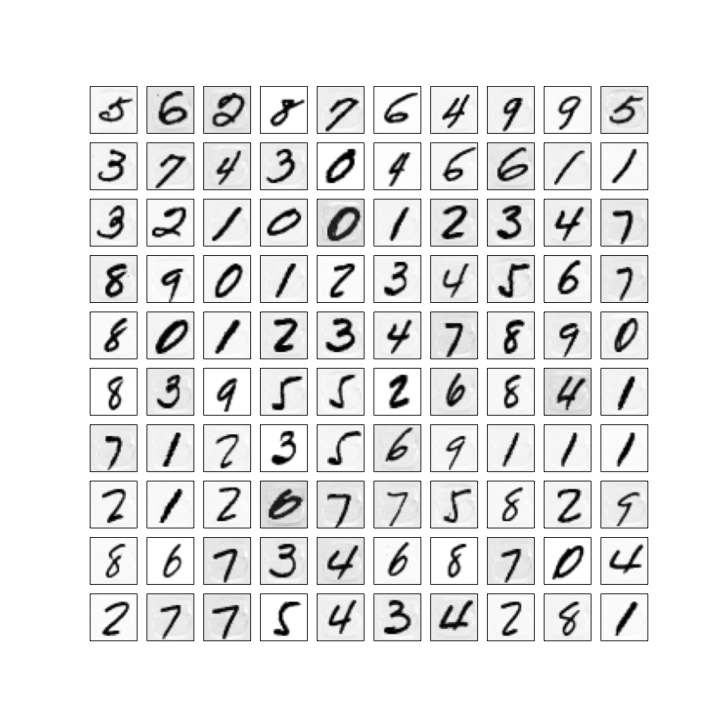}
        \centering{\verb|Attack: EVEN|} \\
%        \centering{\verb|Protect: >=5|}
        \end{minipage}
    \begin{minipage}{0.333\textwidth}
       \includegraphics[trim={2.5cm 2.5cm 2.5cm 3cm},clip,width=\textwidth]{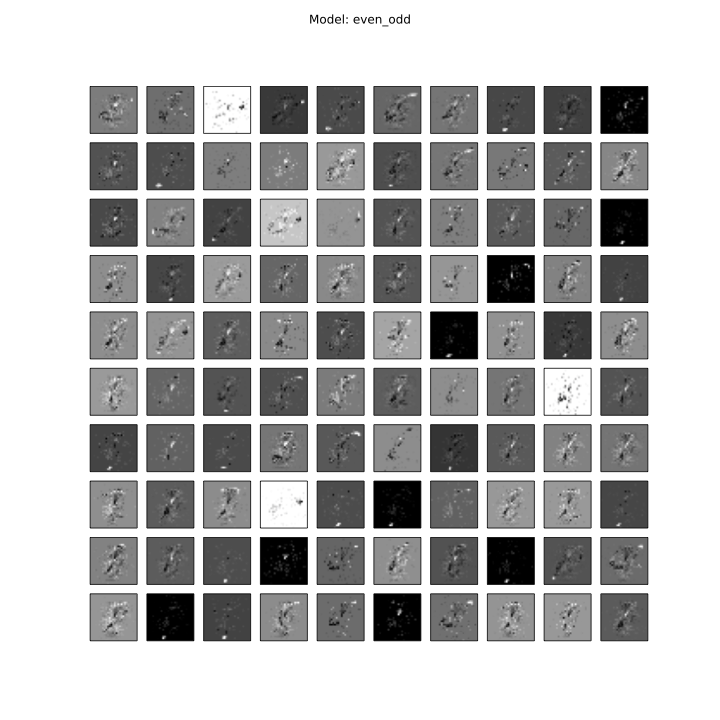}
        \centering{\verb|Exp. EVEN on Original|} \\
%        \centering{\verb|Protect: >=5|}
        \end{minipage}
    \begin{minipage}{0.333\textwidth}
       \includegraphics[trim={2.5cm 2.5cm 2.5cm 3cm},clip,width=\textwidth]{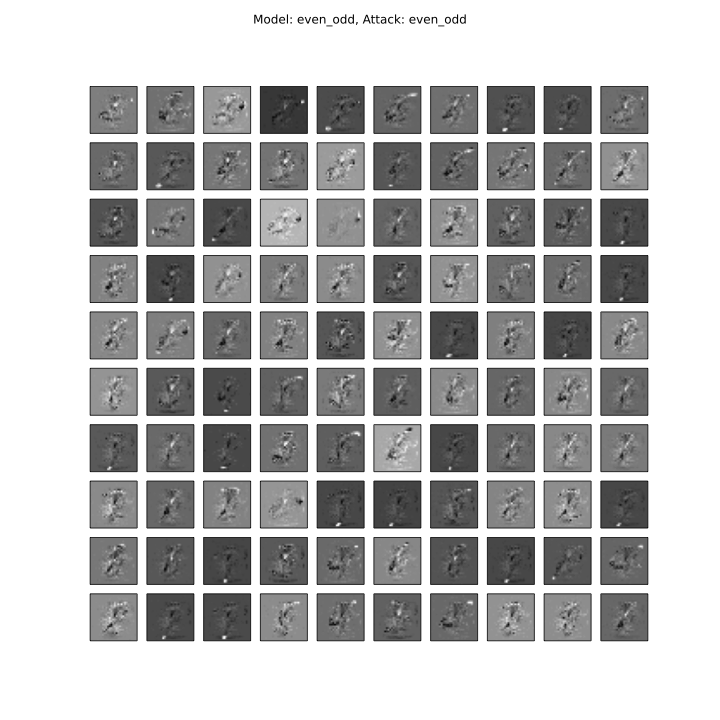}
        \centering{\verb|Exp. EVEN on Attacked|} \\
%        \centering{\verb|Protect: >=5|}
        \end{minipage}
    \caption{\label{fig:mnist_l2_attk1_imgs} Multi-concept attack on the MNIST images with $L_2$ norm minimization. The left plot shows the attacked images, the middle plot shows the SHAP values of the pixels in the original images, and the right plot shows the SHAP values of pixels in the same images after the attack. The predictor  is the "EVEN" classifier.}
\end{figure*}

Figure~\ref{fig:mnist_l2_attk1_imgs} illustrates the results of the $L_2$ norm minimization  when only one concept is attacked. The attacked concept is EVEN and the protected concepts are ZERO and $\ge 5$. The plots show the images of the selected digits after the attack.    

\begin{figure*}[!htb]
    \begin{minipage}{0.245\textwidth}
        \includegraphics[trim={2.5cm 2.5cm 2.5cm 3cm},clip,width=\textwidth]{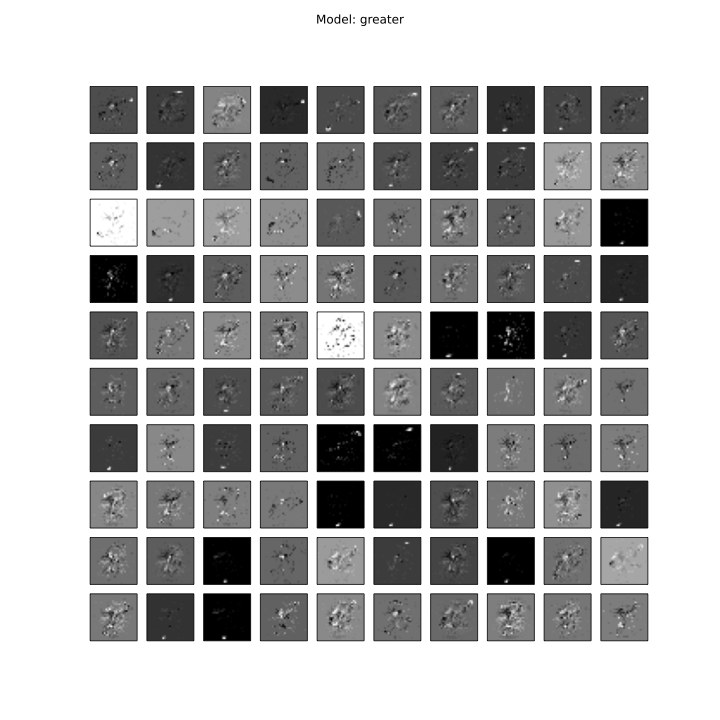}
        \centering{\verb|Exp. >=5 on Original|} 
        \end{minipage}%%
    \begin{minipage}{0.245\textwidth}
        \includegraphics[trim={2.5cm 2.5cm 2.5cm 3cm},clip,width=\textwidth]{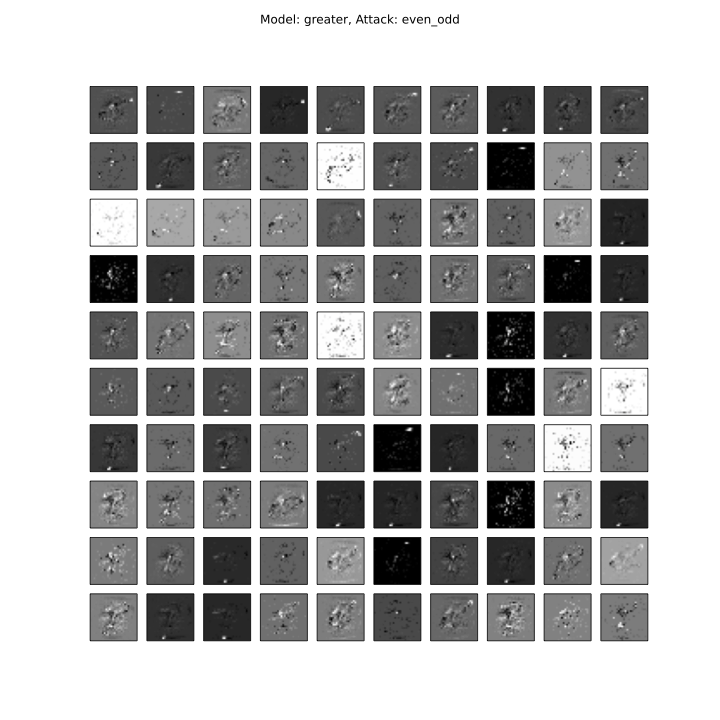}
        \centering{\verb|Exp. >=5 on Attacked|} 
        \end{minipage}%%
    \hfill
    %\centering{\verb|Attack: zero|} 
    %\centering{\verb|Protect: >=5, even number|}
    \begin{minipage}{0.245\textwidth}
        \includegraphics[trim={2.5cm 2.5cm 2.5cm 3cm},clip,width=\textwidth]{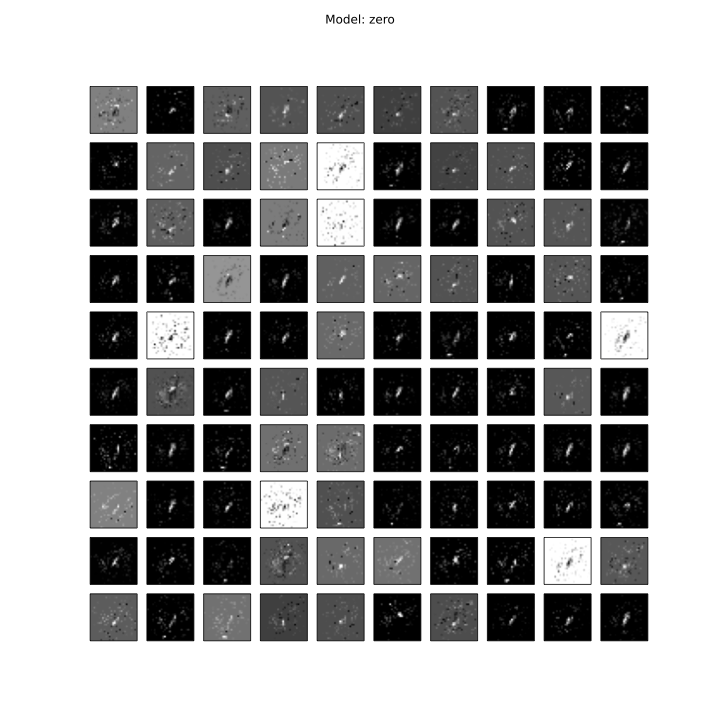}
        \centering{\verb|Exp. ZERO on  Original|} 
        \end{minipage}%%
    \begin{minipage}{0.245\textwidth}
        \includegraphics[trim={2.5cm 2.5cm 2.5cm 3cm},clip,width=\textwidth]{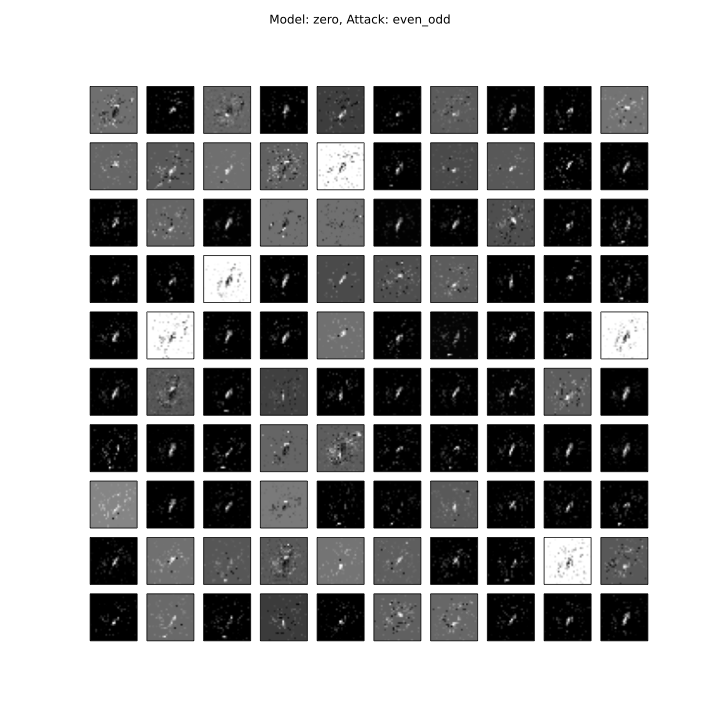}
        \centering{\verb|Exp. ZERO on  Attacked|} 
        \end{minipage}
    \caption{\label{fig:mnist_l2_attk1_protected} Multi-concept attack on the MNIST images with $L_2$ norm minimization. The left plot shows the SHAP values of the pixels in the original images and in the attacked images with the classifier "$\ge 5$" (attacked) as the predictor. The right plot shows the same SHAP values with the classifier ``ZERO'' (also attacked) as the predictor.}
\end{figure*}

Figure~\ref{fig:mnist_l2_attk1_protected} shows the Shapley values of the ``$\ge 5$'' and the ``ZERO'' classifiers, before and after the attack against the ``EVEN'' classifier. The Shapley values do not have significant changes as the two classifiers are protected. 

%*********************************

\begin{figure}[!htb]
    \centering
    \begin{minipage}{0.245\textwidth}
        \includegraphics[width=\textwidth]{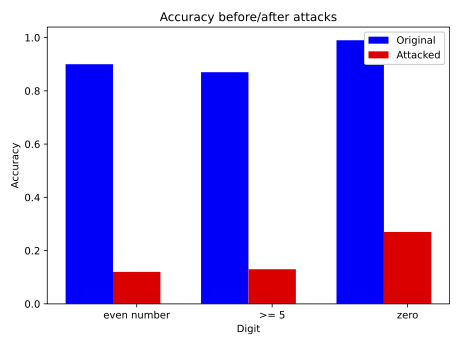}
%        \centering{\verb|Attack: zero|} \\
%        \centering{\verb|Protect: >=5, even number|}
        \end{minipage}%%
    \begin{minipage}{0.245\textwidth}
        \includegraphics[width=\textwidth]{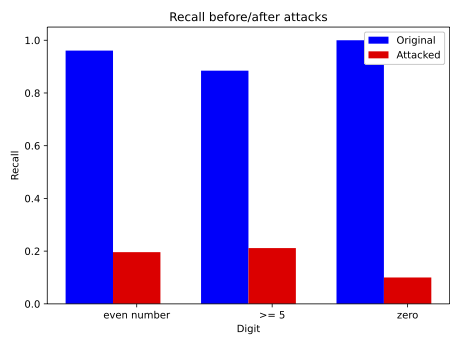}
%        \centering{\verb|Attack: even number, zero|} \\
%        \centering{\verb|Protect: >=5|}
    \end{minipage}
    \caption{\label{fig:mnist_l2_attk3} Multi-concept attack on the MNIST data with $L_2$ norm minimization. All three concepts are attacked.}
\end{figure}

Figure~\ref{fig:mnist_l2_attk3} shows the accuracy and recall when all three classifiers are attacked. The accuracy and recall values of all three classifiers dropped significantly. This  demonstrates the success of simultaneously attacking all classifiers. 
%*****************
% L2 attacked 3
%*****************
    
\begin{figure*}[!h]
    \begin{minipage}{0.333\textwidth}
       \includegraphics[trim={2.5cm 2.5cm 2.5cm 3cm},clip,width=\textwidth]{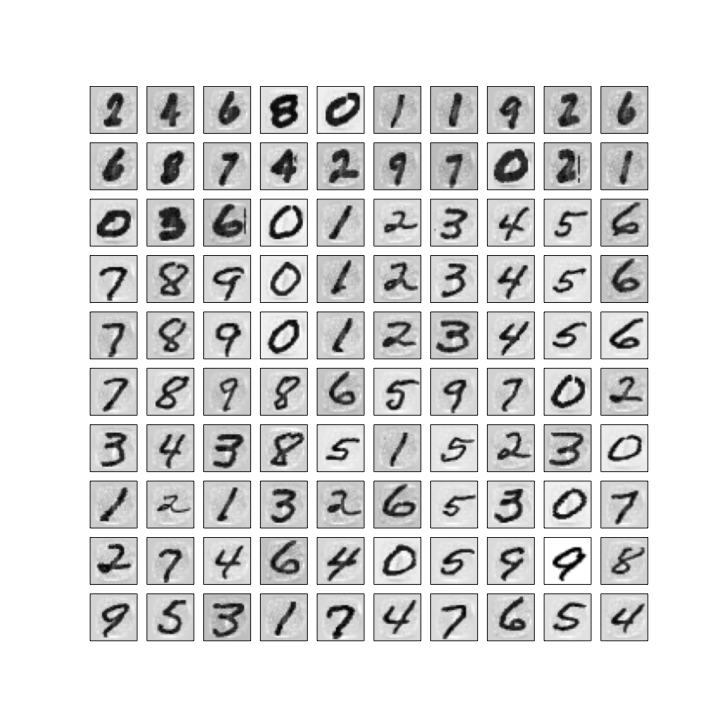}
        \centering{\verb|Attack: EVEN, >=5, ZERO|} \\
%        \centering{\verb|Protect: >=5|}
        \end{minipage}
    \begin{minipage}{0.333\textwidth}
       \includegraphics[trim={2.5cm 2.5cm 2.5cm 3cm},clip,width=\textwidth]{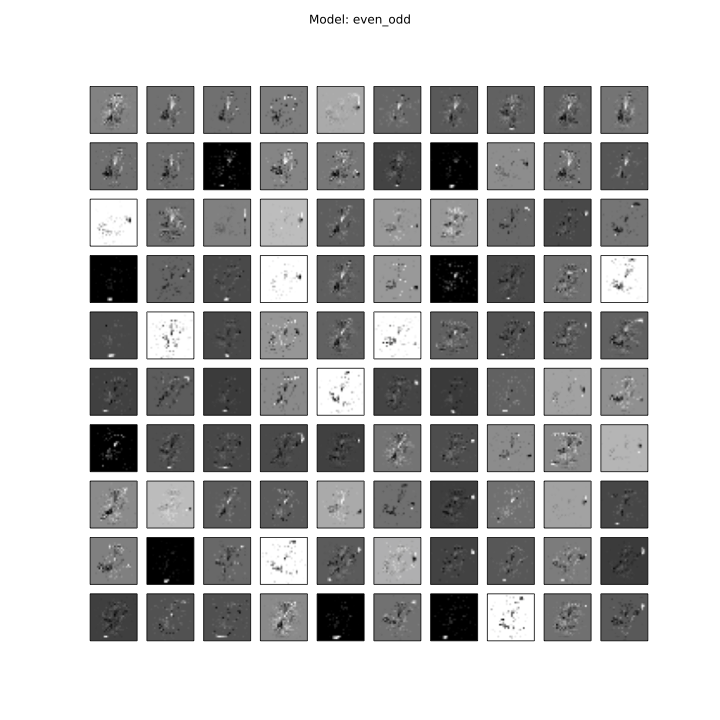}
        \centering{\verb|Exp. EVEN on Original|} \\
%        \centering{\verb|Protect: >=5|}
        \end{minipage}
    \begin{minipage}{0.333\textwidth}
       \includegraphics[trim={2.5cm 2.5cm 2.5cm 3cm},clip,width=\textwidth]{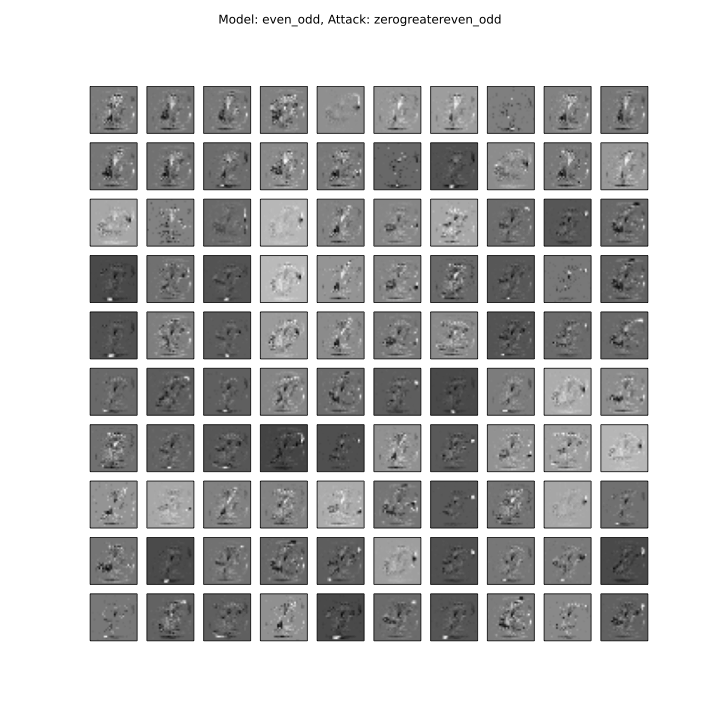}
        \centering{\verb|Exp. EVEN on Attacked|} \\
%        \centering{\verb|Protect: >=5|}
        \end{minipage}
    \caption{\label{fig:mnist_l2_attk3_imgs} Multi-concept attack on the MNIST images with $L_2$ norm minimization. The left plot shows the attacked images, the middle plot shows the SHAP values of the pixels in the original images, and the right plot shows the SHAP values of pixels in the same images after the attack. The predictor  is the "EVEN" classifier.}
\end{figure*}

\begin{figure*}[!htb]
    \begin{minipage}{0.245\textwidth}
        \includegraphics[trim={2.5cm 2.5cm 2.5cm 3cm},clip,width=\textwidth]{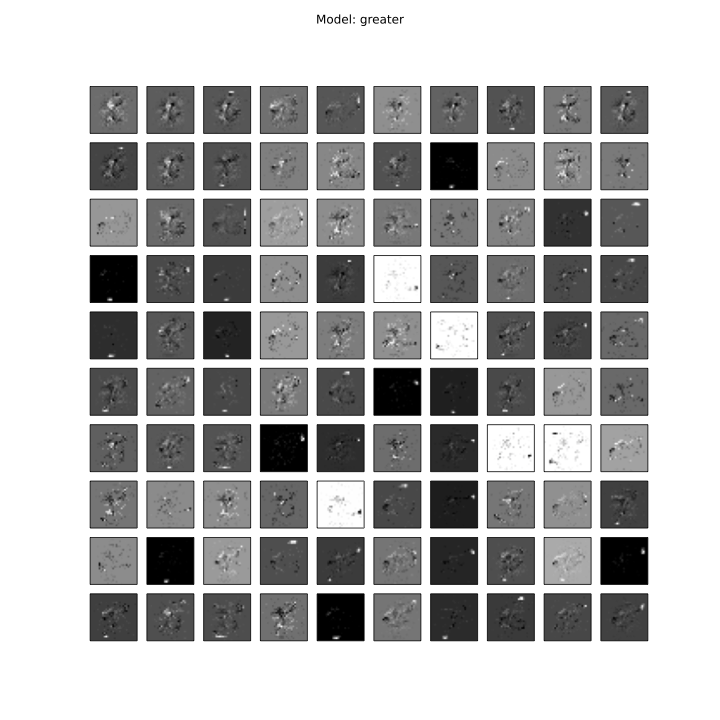}
        \centering{\verb|Exp. >=5 on Original|} 
        \end{minipage}%%
    \begin{minipage}{0.245\textwidth}
        \includegraphics[trim={2.5cm 2.5cm 2.5cm 3cm},clip,width=\textwidth]{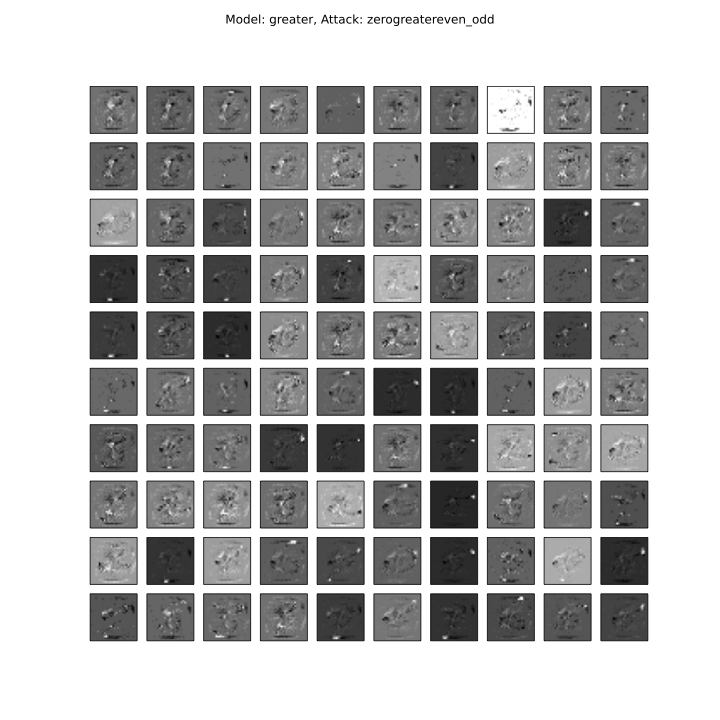}
        \centering{\verb|Exp. >=5 on Attacked|} 
        \end{minipage}%%
    \hfill
    %\centering{\verb|Attack: zero|} 
    %\centering{\verb|Protect: >=5, even number|}
    \begin{minipage}{0.245\textwidth}
        \includegraphics[trim={2.5cm 2.5cm 2.5cm 3cm},clip,width=\textwidth]{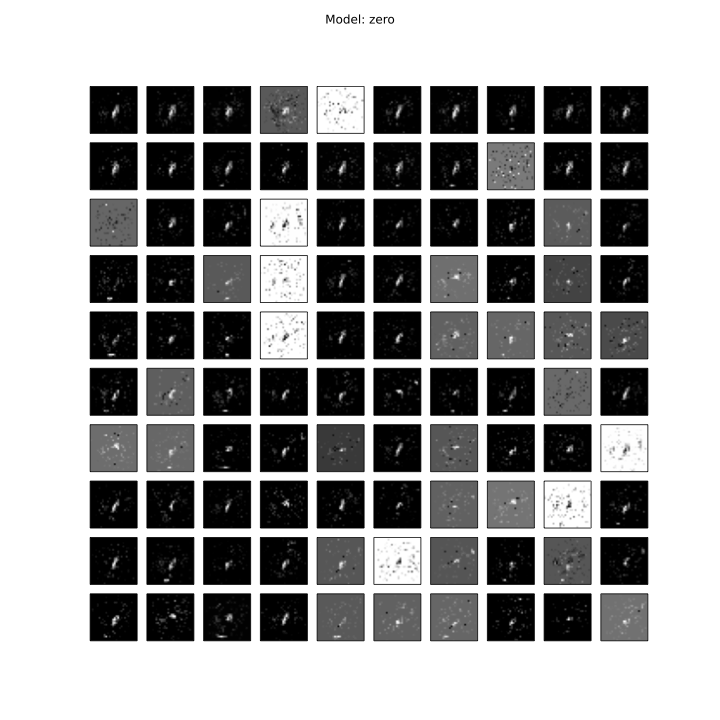}
        \centering{\verb|Exp. ZERO on  Original|} 
        \end{minipage}%%
    \begin{minipage}{0.245\textwidth}
        \includegraphics[trim={2.5cm 2.5cm 2.5cm 3cm},clip,width=\textwidth]{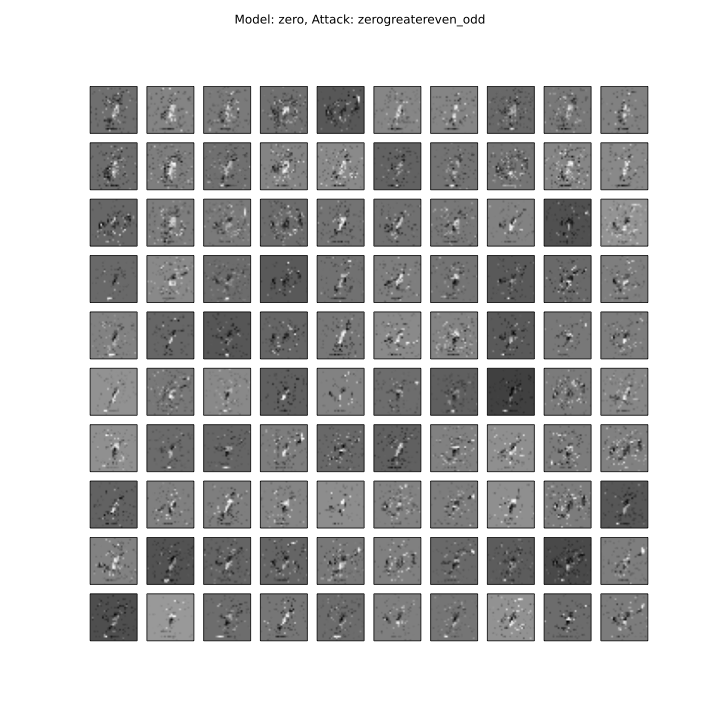}
        \centering{\verb|Exp. ZERO on  Attacked|} 
        \end{minipage}
    \caption{\label{fig:mnist_l2_attk3_protected} Multi-concept attack on the MNIST images with $L_2$ norm minimization. The left plot shows the SHAP values of the pixels in the original images and in the attacked images with the classifier "$\ge 5$" (attacked) as the predictor. The right plot shows the same SHAP values with the classifier ``ZERO'' (also attacked) as the predictor.}
\end{figure*}

Figures~\ref{fig:mnist_l2_attk3_imgs}--~\ref{fig:mnist_l2_attk3_protected} show the Shapley values of all three classifiers before and after the attack. Since all three classifiers are attacked, the Shapley values changed significantly.

\begin{figure}[!htb]
    \centering
    \begin{minipage}{0.245\textwidth}
        \includegraphics[width=\textwidth]{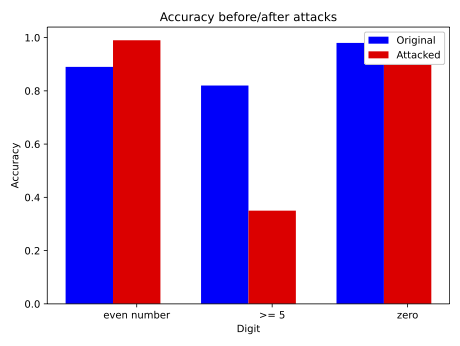}
%        \centering{\verb|Attack: zero|} \\
%        \centering{\verb|Protect: >=5, even number|}
        \end{minipage}%%
    \begin{minipage}{0.245\textwidth}
        \includegraphics[width=\textwidth]{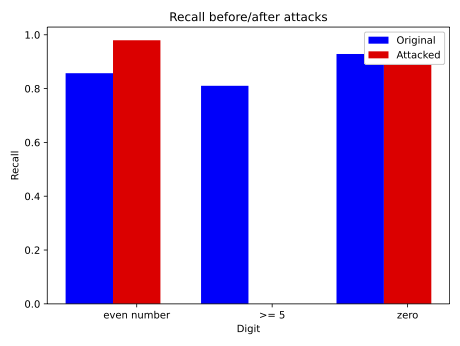}
%        \centering{\verb|Attack: even number, zero|} \\
%        \centering{\verb|Protect: >=5|}
        \end{minipage}
    \caption{\label{fig:mnist_linf_attk1} Multi-concept attack on the MNIST data with $L_\infty$ norm minimization. The concept attacked is {\em >=5} and the protected concepts are {\em zero} and {\em even}.}
\end{figure}

%%%%%%%%%%%%%%%%%%%%%%%%%%%%%%%%%%%%%%%%%%%%%%%%%%%%%%%%%%%%%%%%%%%%%%%%%%%%%%%%
%-------------------------------------------------------------------------------
%*****************
% Linf attacked 1
%*****************
Figure~\ref{fig:mnist_linf_attk1} shows the accuracy and recall of $L_\infty$ norm minimization when only one concept is attacked. The attacked concept is ``>=5'' and and the protected concepts are {\em zero} and {\em even}.  As can be seen, the accuracy and recall of ``>=5'' classifier dropped significantly as the result of attack, while the other two protected classifiers are not affected. 

\begin{figure*}[!h]
    \begin{minipage}{0.333\textwidth}
       \includegraphics[trim={2.5cm 2.5cm 2.5cm 3cm},clip,width=\textwidth]{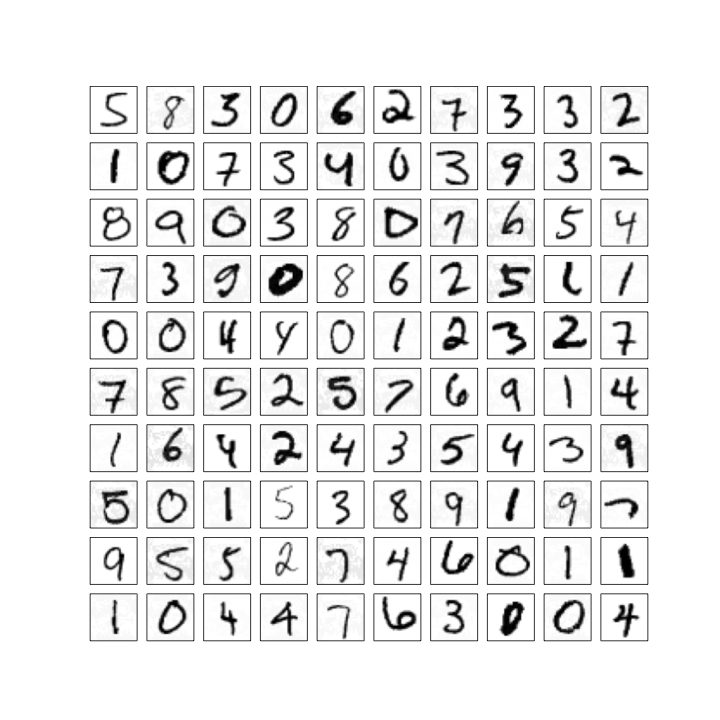}
        \centering{\verb|Attack: >=5|} \\
%        \centering{\verb|Protect: >=5|}
        \end{minipage}
    \begin{minipage}{0.333\textwidth}
       \includegraphics[trim={2.5cm 2.5cm 2.5cm 3cm},clip,width=\textwidth]{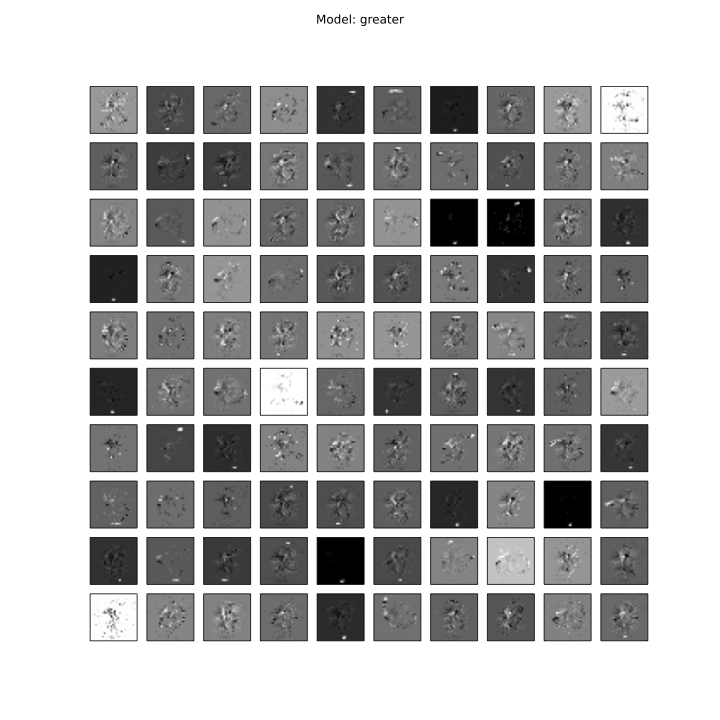}
        \centering{\verb|Exp. >=5 on Original|} \\
%        \centering{\verb|Protect: >=5|}
        \end{minipage}
    \begin{minipage}{0.333\textwidth}
       \includegraphics[trim={2.5cm 2.5cm 2.5cm 3cm},clip,width=\textwidth]{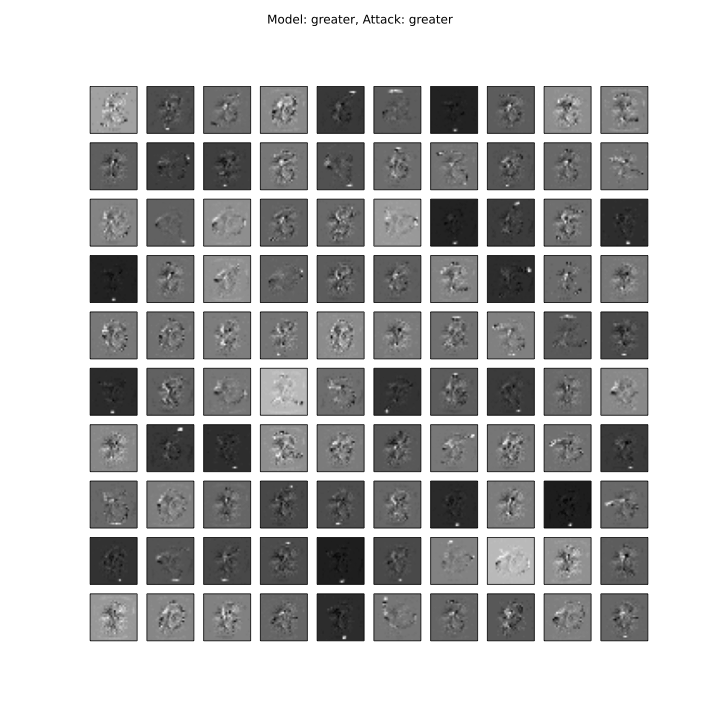}
        \centering{\verb|Exp. >=5 on Attacked|} \\
%        \centering{\verb|Protect: >=5|}
        \end{minipage}
    \caption{\label{fig:mnist_linf_attk1_imgs} Multi-concept attack on the MNIST images with $L_\infty$ norm minimization. The left plot shows the attacked images, the middle plot shows the SHAP values of the pixels in the original images, and the right plot shows the SHAP values of pixels in the same images after the attack. The predictor is the ">=5" classifiers.}
\end{figure*}

Figure~\ref{fig:mnist_linf_attk1_imgs} shows the Shapley values before and after the attack against the ``>=5'' classifier. 

\begin{figure*}[!htb]
    \begin{minipage}{0.245\textwidth}
        \includegraphics[trim={2.5cm 2.5cm 2.5cm 3cm},clip,width=\textwidth]{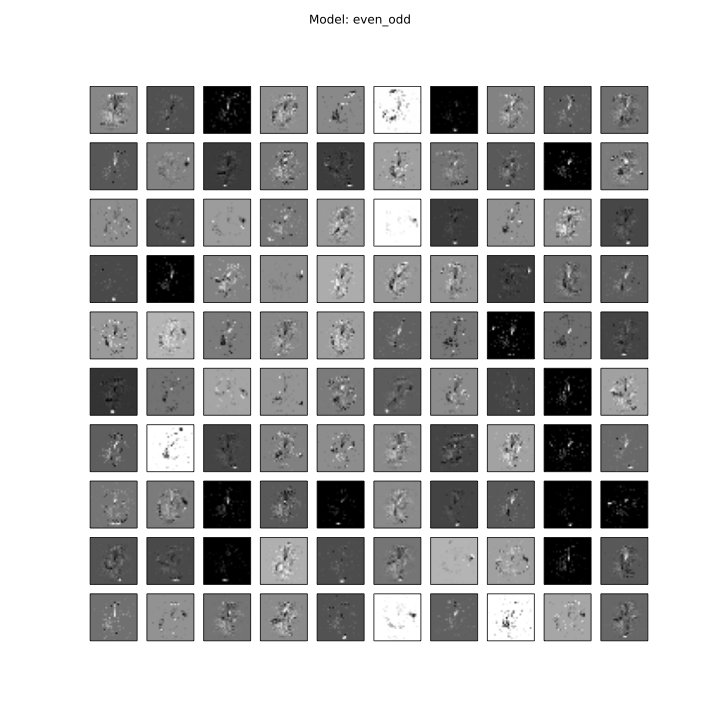}
        \centering{\verb|Exp. EVEN on Original|} 
        \end{minipage}%%
    \begin{minipage}{0.245\textwidth}
        \includegraphics[trim={2.5cm 2.5cm 2.5cm 3cm},clip,width=\textwidth]{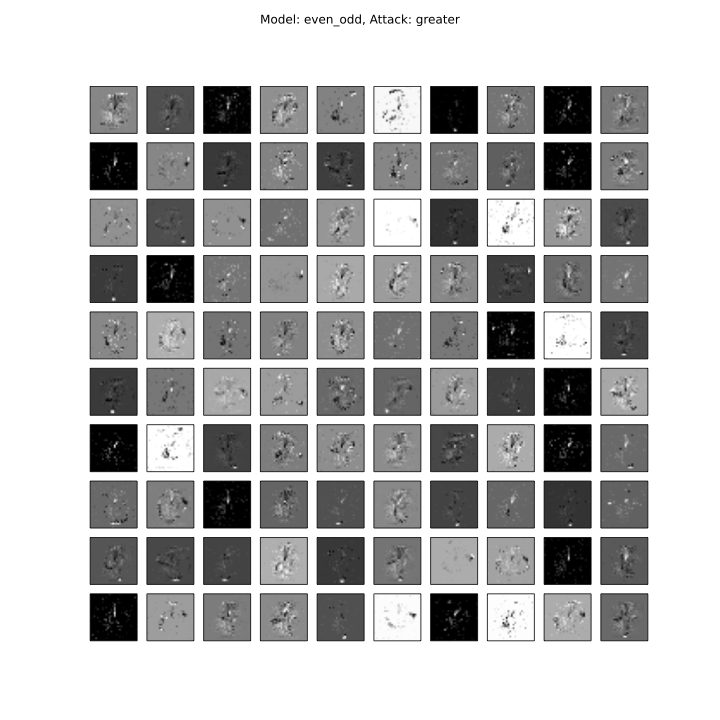}
        \centering{\verb|Exp. EVEN on Attacked|} 
        \end{minipage}%%
    \hfill
    %\centering{\verb|Attack: zero|} 
    %\centering{\verb|Protect: >=5, even number|}
    \begin{minipage}{0.245\textwidth}
        \includegraphics[trim={2.5cm 2.5cm 2.5cm 3cm},clip,width=\textwidth]{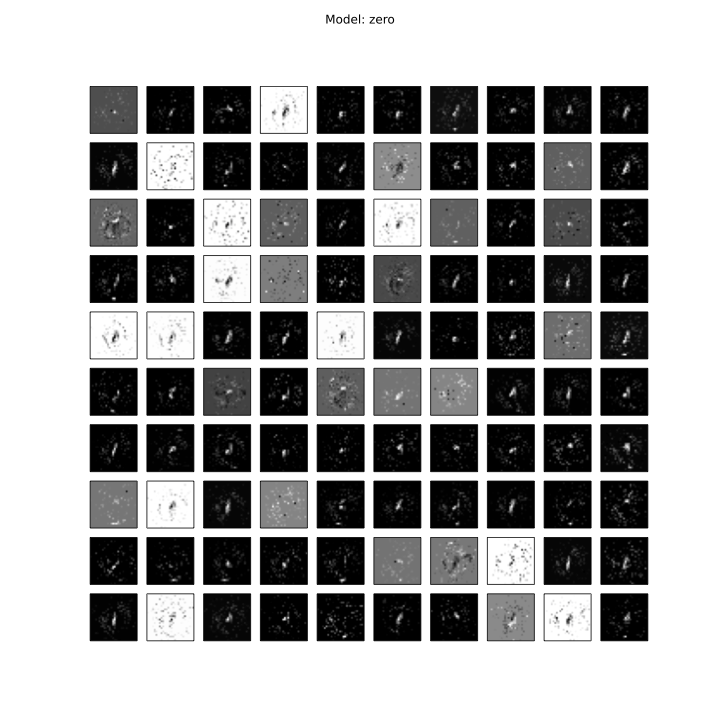}
        \centering{\verb|Exp. ZERO on  Original|} 
        \end{minipage}%%
    \begin{minipage}{0.245\textwidth}
        \includegraphics[trim={2.5cm 2.5cm 2.5cm 3cm},clip,width=\textwidth]{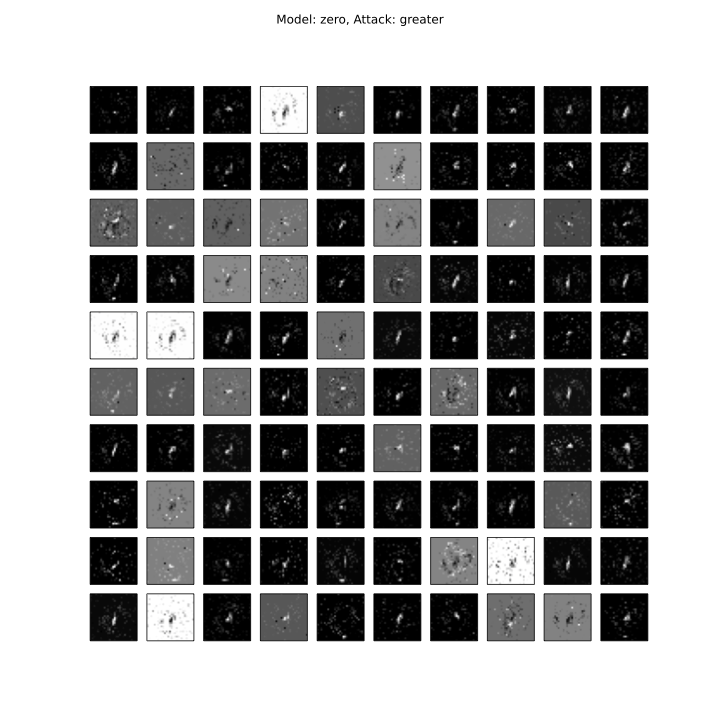}
        \centering{\verb|Exp. ZERO on  Attacked|} 
        \end{minipage}
    \caption{\label{fig:mnist_linf_attk1_protected} Multi-concept attack on the MNIST images with $L_\infty$ norm minimization. The left plot shows the SHAP values of the pixels in the original images and in the attacked images with the classifier "EVEN" (protected) as the predictor. The right plot shows the same SHAP values with the classifier ``ZERO'' (also protected) as the predictor.}
\end{figure*}

Figure~\ref{fig:mnist_linf_attk1_protected} shows the Shapley values of the two protected classifiers ``EVEN'' and ``ZERO'' before and after attack. 

\begin{figure}[!htb]
        \centering
    \begin{minipage}{0.245\textwidth}
    \includegraphics[width=\textwidth]{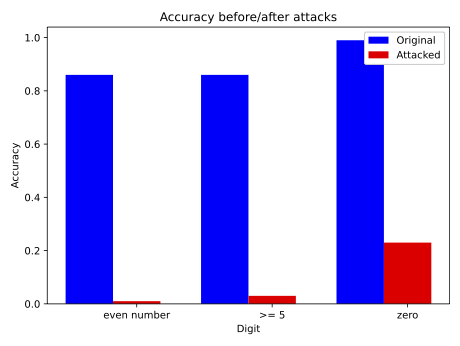}
%        \centering{\verb|Attack: zero|} \\
%        \centering{\verb|Protect: >=5, even number|}
        \end{minipage}%%
    \begin{minipage}{0.245\textwidth}
    \includegraphics[width=\textwidth]{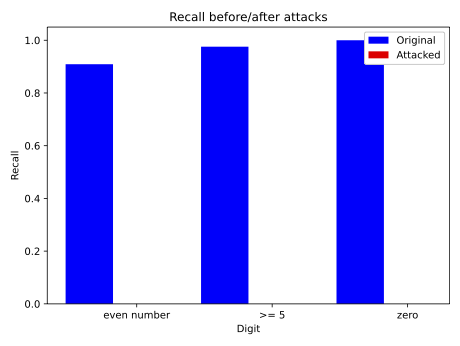}
%        \centering{\verb|Attack: even number, zero|} \\
%        \centering{\verb|Protect: >=5|}
        \end{minipage}
    \caption{\label{fig:mnist_linf_attk3} Multi-concept attack on the MNIST data with $L_\infty$ norm minimization. All three concepts are attacked.}
\end{figure}

%*****************
% Linf attacked 3
%*****************
Figure~\ref{fig:mnist_linf_attk3} shows the accuracy and recall when all three classifiers are attacked. As can be seen, the accuracy and recall values of all three classifiers dropped significantly, including the ``ZERO'' classifier. 

\begin{figure*}[!h]
    \begin{minipage}{0.333\textwidth}
       \includegraphics[trim={2.5cm 2.5cm 2.5cm 3cm},clip,width=\textwidth]{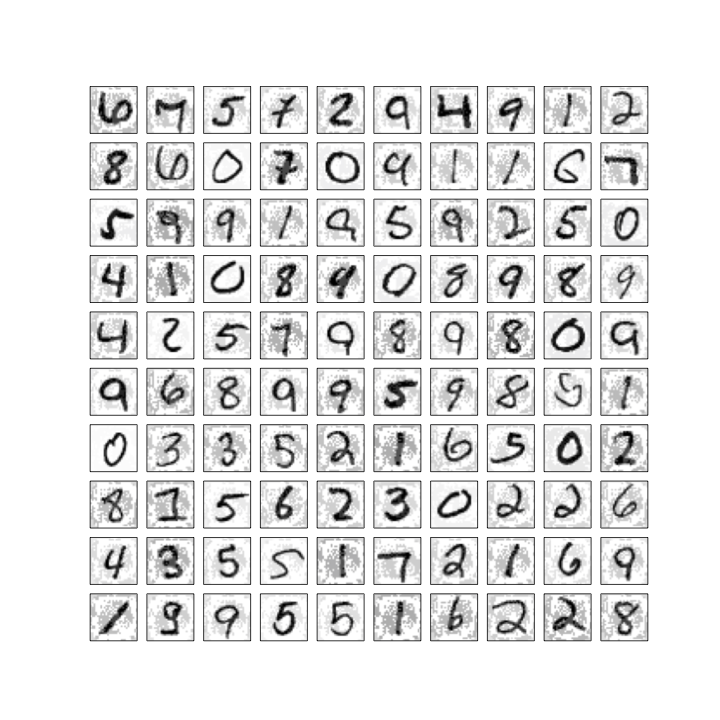}
        \centering{\verb|Attack: >=5, EVEN, ZERO|} \\
%        \centering{\verb|Protect: >=5|}
        \end{minipage}
    \begin{minipage}{0.333\textwidth}
       \includegraphics[trim={2.5cm 2.5cm 2.5cm 3cm},clip,width=\textwidth]{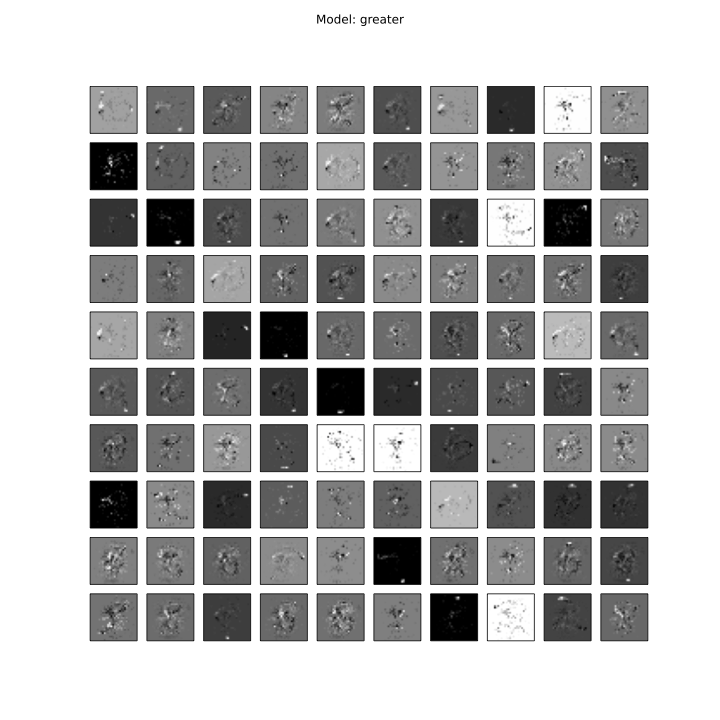}
        \centering{\verb|Exp. >=5 on Original|} \\
%        \centering{\verb|Protect: >=5|}
        \end{minipage}
    \begin{minipage}{0.333\textwidth}
       \includegraphics[trim={2.5cm 2.5cm 2.5cm 3cm},clip,width=\textwidth]{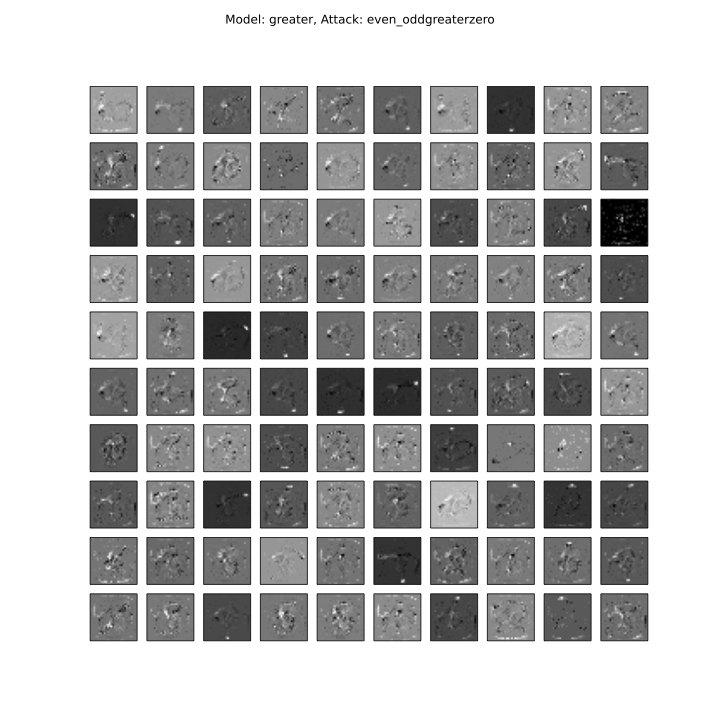}
        \centering{\verb|Exp. >=5 on Attacked|} \\
%        \centering{\verb|Protect: >=5|}
        \end{minipage}
    \caption{\label{fig:mnist_linf_attk3_imgs} Multi-concept attack on the MNIST images with $L_\infty$ norm minimization. The left plot shows the attacked images, the middle plot shows the SHAP values of the pixels in the original images, and the right plot shows the SHAP values of pixels in the same images after the attack. The predictor is the ">=5" classifiers.}
\end{figure*}

\begin{figure*}[!htb]
    \begin{minipage}{0.245\textwidth}
        \includegraphics[trim={2.5cm 2.5cm 2.5cm 3cm},clip,width=\textwidth]{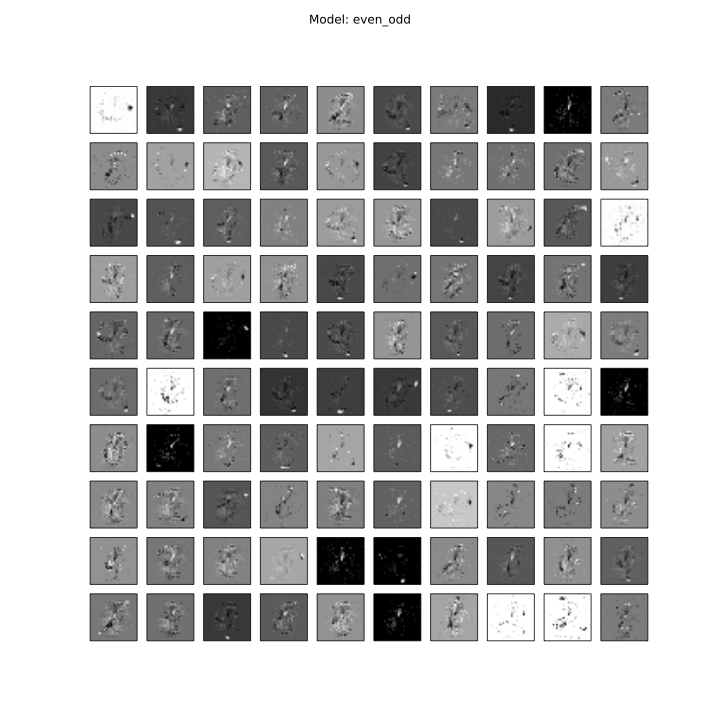}
        \centering{\verb|Exp. EVEN on Original|} 
        \end{minipage}%%
    \begin{minipage}{0.245\textwidth}
        \includegraphics[trim={2.5cm 2.5cm 2.5cm 3cm},clip,width=\textwidth]{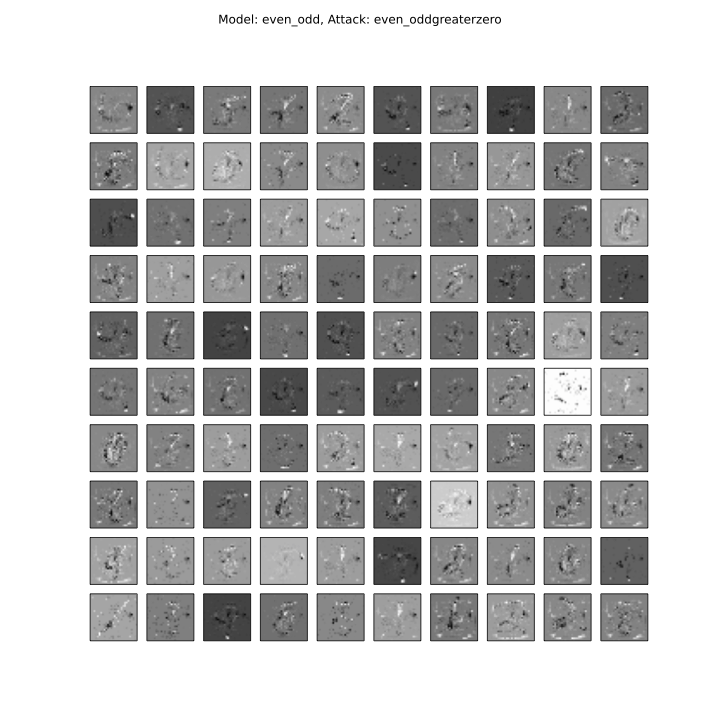}
        \centering{\verb|Exp. EVEN on Attacked|} 
        \end{minipage}%%
        \hfill
        %\centering{\verb|Attack: zero|} 
        %\centering{\verb|Protect: >=5, even number|}
    \begin{minipage}{0.245\textwidth}
        \includegraphics[trim={2.5cm 2.5cm 2.5cm 3cm},clip,width=\textwidth]{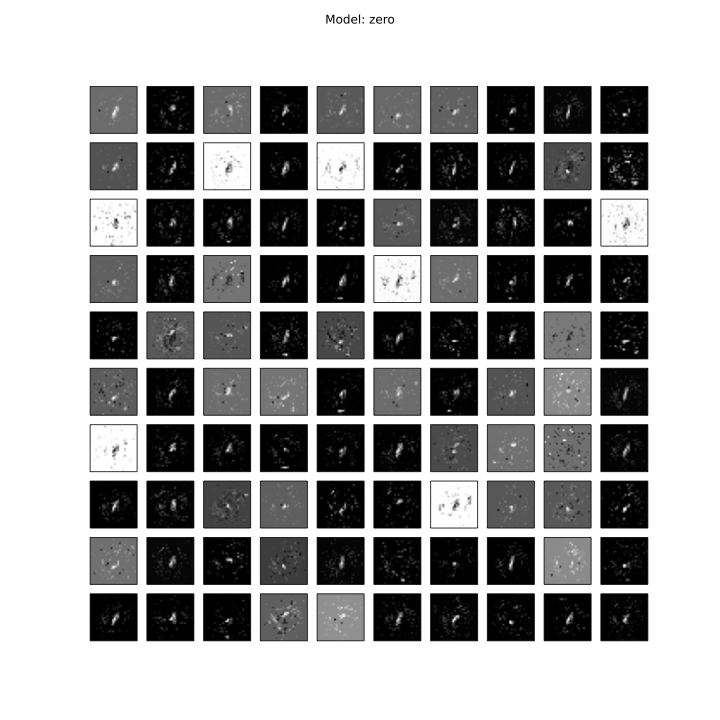}
        \centering{\verb|Exp. EVEN on  Original|} 
        \end{minipage}%%
    \begin{minipage}{0.245\textwidth}
        \includegraphics[trim={2.5cm 2.5cm 2.5cm 3cm},clip,width=\textwidth]{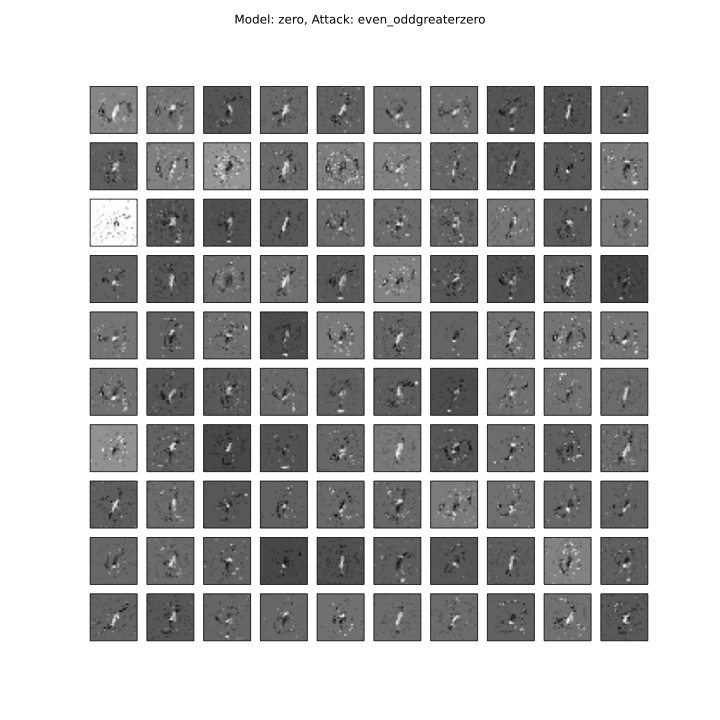}
        \centering{\verb|Exp. EVEN on  Attacked|} 
        \end{minipage}
    \caption{\label{fig:mnist_linf_attk3_protected} Multi-concept attack on the MNIST images with $L_\infty$ norm minimization. The left plot shows the SHAP values of the pixels in the original images and in the attacked images with the classifier "EVEN" (attacked) as the predictor. The right plot shows the same SHAP values with the classifier ``ZERO'' (also attacked) as the predictor.}
\end{figure*}

Figures~\ref{fig:mnist_linf_attk3_imgs}--~\ref{fig:mnist_linf_attk3_protected} show the Shapley value of the three classifiers before and after the attack. 

\end{document}